\documentclass{article}

\usepackage{PRIMEarxiv}

\usepackage[utf8]{inputenc} 
\usepackage[T1]{fontenc}    
\usepackage{hyperref}       
\usepackage{xurl}            
\usepackage{booktabs}       
\usepackage{amsfonts}       
\usepackage{amsmath}        
\usepackage{amssymb}        
\usepackage{nicefrac}       
\usepackage{microtype}      
\usepackage{lipsum}
\usepackage{fancyhdr}       
\usepackage{graphicx}
\usepackage{array}          
\providecommand{\tightlist}{\setlength{\itemsep}{0pt}\setlength{\parskip}{0pt}}
\graphicspath{{./}}         

\title{RAILS:\\ Verification-Native Clearing for Agentic Commerce}

\author{
    Adrian de Valois-Franklin \\
    Evolutionairy AI \\
    Toronto, Canada \\
    \And
    Alex Bogdan \\
    Evolutionairy AI \\
    Toronto, Canada \\
  }

\begin{document}
\maketitle

 \begin{abstract}
Autonomous agents negotiate, purchase, deploy code, and move funds, but no neutral mechanism determines whether they met their delegated obligation, who is responsible when they did not, or which settlement action follows. This is the agentic clearing problem. Tool protocols (MCP), inter-agent communication (A2A), payment rails (x402), mandate and network agent protocols (AP2, Visa, Mastercard), and settlement-risk standards each assume that determination and none produce it.

Clearing is the missing primitive. Payment is not clearing. Authorization is not clearing. LLM-as-judge evaluation is not clearing. Settlement-risk escrow is not clearing: it \emph{consumes} clearing decisions.

RAILS (Real-Time Agent Integrity \& Ledger Settlement) is the integrity and clearing layer for agentic commerce, spanning a per-output reliability score, a published reliability record, and a clearing function that consumes them. The clearing protocol at its core closes that gap. Seven primitives (Obligation Object, Evidence Envelope, Verification Mesh, Clearing Decision, Settlement Instruction, Clearing Passport, Finality Rules), bound by a formal model of admissibility-graded verification, together yield a soundness property: no financially material settlement is supported by evidence below the obligation's admissibility floor. The property is falsifiable against the spec. We are not aware of a prior agent-commerce verification mechanism that states a property of this kind. The approaches nearest to it emit a pass, a delivery guarantee, a bare score, or an equilibrium.

This paper specifies that clearing protocol.

\end{abstract}


\section{The Agentic Clearing Problem}

A coding agent commits a patch to a production database service. A procurement agent reads a spec, decides a \$40{,}000 part qualifies, and releases the payment. An enterprise orchestrator hires a legal agent to draft a supply contract, signed before anyone catches the indemnity it waived. A travel agent books nonrefundable flights for the wrong week. None of these needed a human in the loop, and each consequence dwarfs the model call that produced it. What is new is not that the actions are consequential. It is that the consequence settles before anything verifies the judgment behind it.

\textbf{The unit of concern in artificial intelligence is no longer the generated sentence. It is the delegated action.}

An agent can hold valid authorization, settle a valid payment, to an honest merchant who delivers exactly what was asked, and still leave the user harmed, because the agent asked for the wrong thing. Performance failure is not counterparty failure, and no payment or authorization rail is built to catch it. That gap is what this paper addresses.

A growing infrastructure has emerged around this shift. Tool protocols such as Anthropic's Model Context Protocol (MCP)~\cite{mcp} connect agents to external systems. Agent communication protocols such as Google's Agent2Agent (A2A)~\cite{a2a} let agents discover and coordinate with one another. Mandate frameworks such as Google's Agent Payments Protocol (AP2)~\cite{ap2} bind user authorization to bounded actions. Payment rails such as Coinbase's x402~\cite{x402} execute machine-native value transfer. Payment networks add their own agent layers: Visa's Trusted Agent Protocol and Intelligent Commerce~\cite{visa_tap}, and Mastercard's Agent Pay and Verifiable Intent~\cite{mastercard_vi}, all of which attest agent identity and user authorization at the point of transaction. Settlement-risk standards escrow fees, post collateral, and pay claims when covered failures occur. Every one of these layers is necessary; none is sufficient.

They are not sufficient because they all rest on a determination none of them produce. A tool protocol records what was called; it does not say whether the call satisfied the obligation. A payment rail executes a transfer; it does not say whether the service was delivered. A settlement-risk standard releases a fee on a covered outcome; the outcome itself must be determined by something else. That something else is what we call the \textbf{agentic clearing problem}.

The verification work nearest to this problem gates settlement on a check of execution, but what it emits is a boolean pass, a delivery guarantee, a numeric score, or a game-theoretic equilibrium. None of these is a determination graded by the admissibility of its evidence, and none states a property that can be falsified against a specification. That is the gap RAILS closes.

\subsection*{Six separable questions}

A transaction involving an autonomous agent contains six separable questions. They are not interchangeable; they cannot be answered by a single signal; and any attempt to settle the transaction without answering all of them either ignores risk or imposes it on the wrong party.

\begin{itemize}
\tightlist
\item \emph{Authorization} asks whether the agent was allowed to act.
\item \emph{Execution} asks what the agent actually did.
\item \emph{Performance} asks whether the action satisfied the delegated obligation.
\item \emph{Attribution} asks, if something failed, who or what caused it.
\item \emph{Loss} asks what material, operational, or contractual harm occurred.
\item \emph{Settlement} asks what financial, reputational, or procedural consequence follows.
\end{itemize}

Current infrastructure answers these in fragments: authorization protocols handle the first, tool logs the second, benchmarks the third, security tools the fourth, insurance frameworks the fifth and sixth. The pieces exist. The lifecycle that binds them does not.

\subsection*{What clearing is not}

\textbf{Why authorization is not clearing.} A human can authorize a travel agent to book a flight; the agent may book the wrong airport. A user can authorize a coding agent to modify a repository; the agent may introduce a vulnerability. A buyer can authorize a procurement agent to purchase from a merchant; the merchant may substitute a noncompliant item. Authorization establishes permitted agency. Clearing determines fulfilled obligation.

\textbf{Why payment is not clearing.} Payment rails establish transfer. They do not establish performance. An HTTP-native payment can be cryptographically valid and still correspond to a defective service. A card authorization can be valid and still result in a chargeback. A stablecoin transfer can settle on-chain and still leave an off-chain service undelivered. Payment settles value transfer. Clearing settles obligation state.

\textbf{Why escrow and underwriting are not clearing.} Escrow delays payment until conditions are met. Underwriting compensates for covered failures. Both depend on a prior mechanism for determining whether the condition was met or the covered failure occurred. Without clearing, the system relies on self-reporting, human arbitration, an opaque judge model, or bespoke domain logic. Escrow and underwriting are settlement mechanisms. Clearing is what produces the decision they execute.

\textbf{Why LLM-as-judge is not clearing.} LLM judges can interpret semantic performance, especially for open-ended deliverables. But they suffer from position bias, verbosity bias, self-enhancement bias, domain fragility, prompt sensitivity, and vulnerability to adversarial framing \cite{zheng2023judging}. In a clearing context, these limitations are amplified: a judge does not merely rank outputs; it may determine fee release, collateral slash, claim eligibility, liability assignment, or reputation damage. A judge is one signal. Clearing is a governed decision process.

\subsection*{The composite question}

The agentic clearing problem can be stated as a single composite question:

\begin{quote}
\emph{What was promised, what happened, what evidence is admissible, who is responsible, what loss occurred, and what settlement action is final?}
\end{quote}

This is the question every risk-bearing settlement system presupposes, and that no authorization, communication, or payment protocol answers. The verification systems that do attempt an answer emit a pass, a delivery guarantee, a score, or an equilibrium, not the neutral, admissibility-graded clearing decision a sound settlement requires. A tool protocol does not produce it. An inter-agent communication protocol does not produce it. A payment rail does not produce it. A risk-standard escrow state machine does not produce it. Even an LLM judge does not produce it: a judge gives an opinion, not a clearing decision.

The composite question is also not a research benchmark. A benchmark~\cite{jimenez2023swebench, liu2023agentbench, wang2024battleagent} says whether a model solved an issue in a test environment. Clearing says whether a real agent satisfied a real obligation, whether the evidence is admissible, whether settlement should release funds, and whether the agent remains eligible for future clearing. Benchmarks measure capability. Clearing measures fulfillment.

\subsection*{RAILS}

RAILS, a verification-native clearinghouse protocol, is the answer this paper develops. It consists of seven primitives. An \emph{Obligation Object} compiles natural-language intent into a signed, machine-clearable contract. An \emph{Evidence Envelope} binds execution artifacts to that obligation under explicit admissibility classes. A \emph{Verification Mesh} applies multiple verifier types (receipt verifiers, constraint verifiers, semantic verifiers, human arbiters, and others) under conflict-resolution and escalation rules. A \emph{Clearing Decision} aggregates the mesh's outputs into a structured verdict. A \emph{Settlement Instruction} translates that verdict into downstream actions. A \emph{Clearing Passport} records cross-transaction reliability for agents, tools, and providers. \emph{Finality Rules} gate when settlement becomes binding.

The formal contribution is a property the protocol enforces. Verifiers must declare the evidence basis they relied on. The mesh aggregator weights verifier votes by the admissibility class of their bases. No financially material settlement instruction may carry a basis below the obligation's admissibility floor. This is the protocol's soundness property: unlike a manifesto's promise, it is falsifiable against the specification.

RAILS is positioned among existing protocols, not against them. It consumes tool traces from MCP and subdelegation events from A2A as evidence inputs. It imports AP2-style mandates as authority evidence. It targets x402 endpoints, escrow systems, and enterprise ledgers as settlement rails. Settlement-risk standards, in this layering, are \emph{consumers} of the Settlement Instruction: their settlement state transitions correspond to RAILS settlement subtypes (Section 9.4).

\subsection*{Thesis}

The central claim of this paper can be stated in one sentence.

\begin{quote}
\emph{Autonomous-agent commerce cannot scale without a neutral mechanism that converts execution traces into admissible evidence, aggregates verifier outputs under explicit admissibility classes, and emits settlement instructions that are sound --- meaning no financially material settlement is ever supported by inadmissible evidence. RAILS is that mechanism.}
\end{quote}

The rest of this paper exists to make that claim precise enough to falsify.

\section{Related Work}

A layer of protocols now standardizes how agents call tools, coordinate, carry authorization, and move value. Tool protocols such as MCP expose typed tool traces, agent-communication protocols such as A2A handle discovery and subdelegation, mandate and commerce protocols such as AP2, UCP~\cite{ucp}, and the OpenAI and Stripe Agentic Commerce Protocol (ACP)~\cite{acp_openai} bind user authorization to bounded actions, and payment rails such as x402 execute machine-native transfer. Payment networks add their own agent layers in the same band: Visa's Trusted Agent Protocol and Intelligent Commerce, and Mastercard's Agent Pay and Verifiable Intent, which attest agent identity and user authorization at checkout. Each standardizes a part of the lifecycle. None determines whether the agent's action satisfied the obligation. A recent systematization of agent-to-agent payments states the gap directly: payment completion is only weakly coupled to service completion, and financial finality does not establish fulfillment~\cite{sok_a2a_2026}.

A second line of work specifies what happens to money after an obligation runs. Settlement-risk standards, such as~\cite{hua2026ars}, organize escrow lock and release, collateral posting, underwriting, claims, and reimbursement into a settlement state machine. These standards provide settlement discipline, yet every financially material transition consumes a prior determination of whether the obligation was met, a determination they leave external and do not produce. RAILS supplies the determination they presuppose; Section 9.4 develops the consumer relationship in detail.

Closest to RAILS is recent work that gates payment on a check of execution. Verify-then-pay infrastructure releases funds when an execution proof satisfies fixed verification predicates~\cite{tesspay2026}. Enforcement-based coupling finalizes a transfer if and only if a service is correctly executed and delivered, using trusted-execution adaptor signatures~\cite{a402_2026}. Trustless-verification protocols make correctness an equilibrium by letting challengers stake to expose error, slashing incorrect agents and erroneous verifiers along an escalation path~\cite{operator2025}. An on-chain commerce protocol in the same family adds an explicit evaluation phase that gates escrow release through a fee-earning evaluator agent. These approaches share the premise that settlement must follow verification. They differ from RAILS in what they emit: a boolean pass, a delivery guarantee, a numeric score, or a game-theoretic equilibrium, rather than a determination graded by the admissibility of its evidence and carrying a property that can be falsified against a specification.

A parallel ecosystem evaluates agent and model behavior. LLM-as-judge methods interpret open-ended performance but exhibit position, verbosity, and self-enhancement bias and are vulnerable to adversarial framing~\cite{zheng2023judging}. Surveys of agent-as-judge evaluation and of inter-agent trust models reach a compatible conclusion, that proofs guarantee integrity rather than alignment and that judging can complement but not replace governed oversight~\cite{hu2025trustmodels}. On-chain standards such as ERC-8004 provide identity, reputation, and validation registries for agents~\cite{erc8004}, a substrate on which a validation function can record results, while fixing neither the validation method nor any property the result must satisfy.

Against this landscape RAILS makes three claims the surveyed work does not jointly make. The determination is graded by the admissibility of its evidence rather than reduced to a pass or a score. A soundness property guarantees that no financially material settlement rests on evidence below the obligation's admissibility floor, and the property is falsifiable against the specification. The clearing function is neutral, neither a fee-earning market participant nor a self-interested validator set. The remainder of the paper specifies the protocol that carries these properties.

\section{RAILS at a Glance}

\begin{figure}
\centering
\includegraphics[width=\linewidth,keepaspectratio]{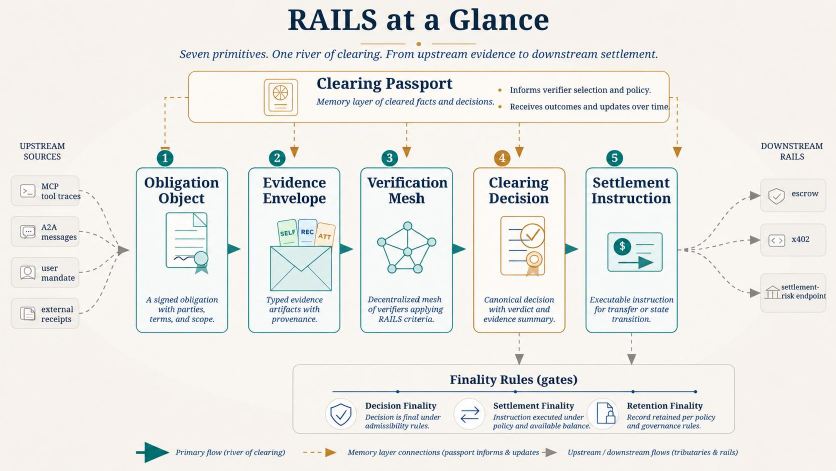}
\caption{RAILS at a glance. Seven primitives convert signed intent and execution evidence into a settleable decision. Five sit along the main flow (Obligation Object, Evidence Envelope, Verification Mesh, Clearing Decision, Settlement Instruction); the Clearing Passport (above) supplies cross-transaction reliability to verifier selection and absorbs each clearing outcome; the Finality Rules (below) gate when a decision becomes binding. Upstream protocols (MCP, A2A, user mandates, external receipts) feed evidence in; downstream rails (escrow, x402, settlement-risk endpoints) execute the resulting settlement instruction.}
\label{fig:rails-glance}
\end{figure}

RAILS (Real-Time Agent Integrity \& Ledger Settlement) is a verification-native clearinghouse protocol for autonomous-agent obligations. Figure~\ref{fig:rails-glance} shows the protocol at a glance. The input is signed intent (an Obligation Object) together with the post-execution evidence captured into an Evidence Envelope. The output is a Clearing Decision (the verdict the parties needed) and a Settlement Instruction (the action the settlement systems should execute). Between the two sit five further primitives, a ten-stage lifecycle, and a single architectural claim: that clearing is the layer the agentic stack has been missing.

\subsection*{The seven primitives}

The \textbf{Obligation Object} is the contract. It compiles natural-language intent into a signed, machine-clearable structure: parties and agents, task statement and scope, required and prohibited actions, acceptance criteria, evidence requirements, admissibility floor, settlement policy, finality policy. The contracting parties produce it at binding; every downstream primitive consumes it as the canonical reference. Without an Obligation Object, there is nothing to clear.

The \textbf{Evidence Envelope} is the record of what happened. It is a hash-anchored container of evidence items captured during execution (git diffs, CI logs, scanner outputs, third-party receipts, reviewer comments, agent self-reports); each item carries its provenance chain and its admissibility tag. The executing agent assembles it, with third-party evidence emitters (CI runners, scanners, reviewers) writing in directly. The Verification Mesh consumes it as its evidentiary input.

The \textbf{Verification Mesh} is the panel. It is a configured set of heterogeneous verifiers (constraint checkers, receipt verifiers, semantic judges, policy checkers, human arbiters), each of which returns a verdict, a confidence, and a declared evidence basis. The obligation's verifier policy determines which verifiers run for any given clearing event. Mesh members produce their outputs independently; the Clearing Engine consumes them as the raw material of the Clearing Decision.

The \textbf{Clearing Decision} is the canonical output of the protocol. It is a structured verdict carrying a performance status (did the task succeed?), a policy status (did the agent follow the rules?), a fault assignment, a loss estimate, the evidence basis the verdict rests on, an aggregated confidence, and a finality status. The Clearing Engine (the aggregator function that combines mesh outputs under the obligation's admissibility floor) produces it. Both the Settlement Instruction and the Clearing Passport consume it. The protocol's soundness guarantee lives here: a Clearing Decision with consequence cannot rest on evidence weaker than the floor the obligation declared.

The \textbf{Settlement Instruction} is the action. It is a structured action list (\texttt{fee\_action}, \texttt{principal\_action}, \texttt{collateral\_action}, \texttt{claim\_action}, \texttt{penalty\_action}, \texttt{reputation\_action}), targeted at a named execution rail. The Clearing Decision and the obligation's settlement policy together produce it. Downstream rails consume it: x402 endpoints, escrow systems, enterprise ledgers, settlement-risk standards. Settlement-risk standards, in this layering, are consumers of the Settlement Instruction; Section 9.4 develops that point.

The \textbf{Clearing Passport} is the memory. It is a cross-transaction reliability record for agents, tools, providers, and verifiers: what they have cleared, what has been disputed, where their declared bases have drifted from their actual performance. The Clearing Engine updates it on each emitted decision. Future obligations consume it during pre-clearance to inform verifier selection, exposure scoring, and confidence priors. Passports are how the protocol learns across transactions.

The Passport is private and per-agent. The same reliability-scoring machinery, applied at model scope rather than agent scope and published openly, yields independent integrity ratings on frontier models, RAILS Report Cards. Report Cards are to models what Passports are to agents: a continuously updated, comparable reliability signal. In protocol terms they are one natural source of the verifier reliability priors and confidence priors the aggregator and Pre-Clearance consume; a verifier built on a given model inherits a prior informed by that model's public rating. The private Passport and the public Report Card are the same scoring at two scopes.

\textbf{Finality Rules} are the gate. They form a predicate (admissibility floor met, confidence threshold met, conflict resolution complete, appeal window expired) whose firing transitions a Settlement Instruction from PROVISIONAL to FINAL. The obligation's finality policy produces the rules. The rails executing the instruction consume them, holding reversible settlement open until the predicate is satisfied. PROVISIONAL is the protocol's tolerance for being wrong; FINAL is its commitment to the outcome.

\subsection*{Lifecycle in four phases}

The lifecycle in which these primitives operate runs from discovery to finality in ten stages and falls naturally into four phases. \emph{Binding} (discovery, negotiation, obligation binding) is where intent becomes a signed contract; this phase is reversible, since parties can renegotiate until the Obligation Object is signed. \emph{Execution and evidence} (pre-clearance, execution, evidence submission) is where the agent acts and the artifacts of its action are captured into the Envelope; this phase can be paused, for example to recompute margin when exposure shifts mid-task. \emph{Verification and clearing} (verification, adjudication) is where the Mesh runs and the Clearing Engine aggregates its outputs into a Clearing Decision. \emph{Settlement and finality} (settlement instruction, finality) closes the loop: the Settlement Instruction is emitted, executed through the chosen rail, and held open in PROVISIONAL state until the Finality Rules predicate fires.

Finality is the point past which a Clearing Decision becomes binding. Section 6 develops the lifecycle as an explicit state machine (Figure~\ref{fig:lifecycle}).

\subsection*{Where RAILS sits in the stack}

\begin{figure}
\centering
\includegraphics[width=\linewidth,height=0.72\textheight,keepaspectratio]{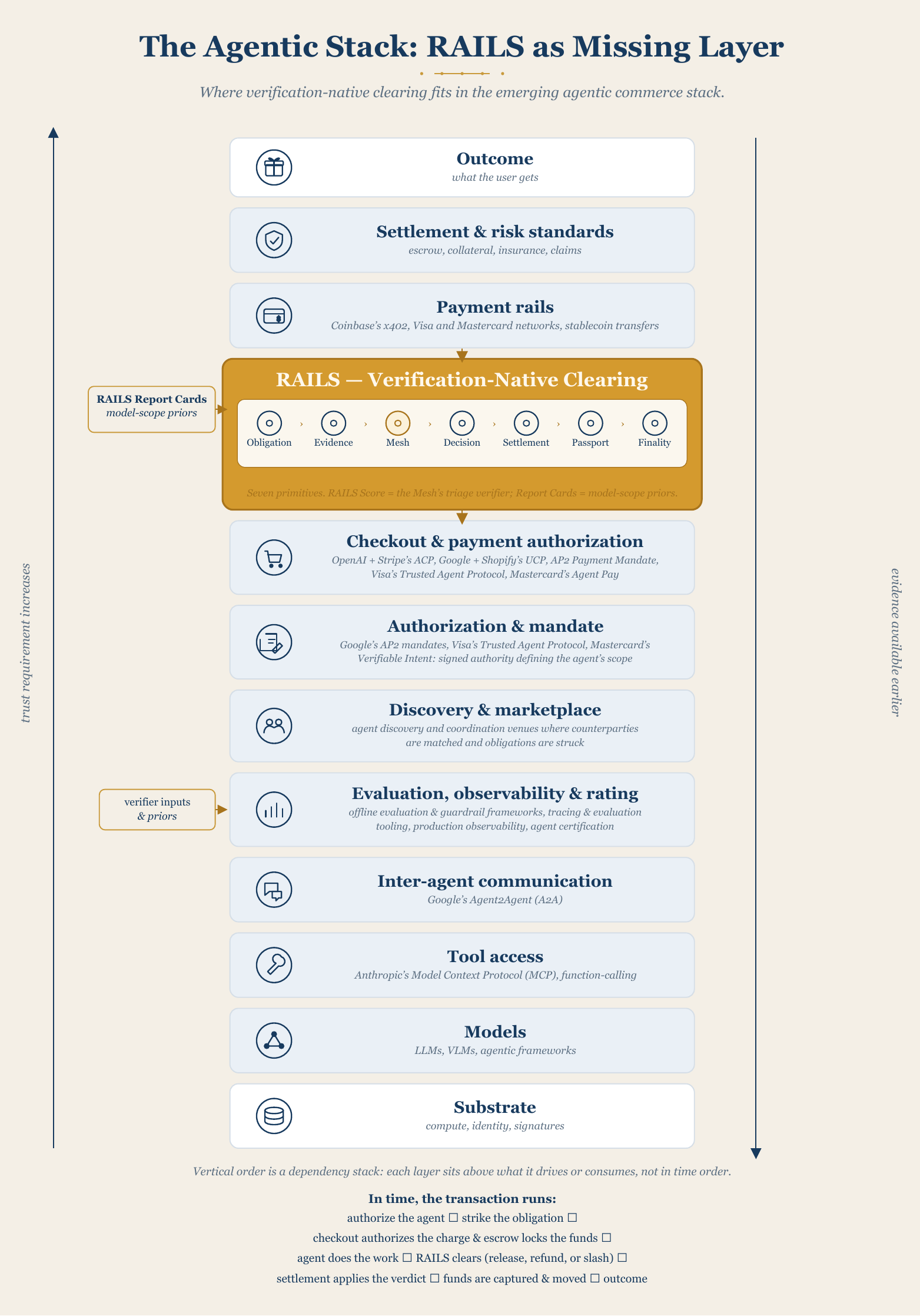}
\caption{The agentic stack with RAILS as the missing clearing layer. Existing infrastructure has converged on models; tool access (Anthropic's MCP); inter-agent communication (Google's A2A); a discovery-and-marketplace layer where counterparties are matched and obligations struck; authorization, mandate, and checkout (Google's AP2, Google and Shopify's UCP, OpenAI and Stripe's ACP, and Visa's and Mastercard's agent protocols); payment rails (Coinbase's x402); settlement-risk standards (escrow, collateral, claims); and a parallel evaluation, observability, and rating ecosystem. Each layer assumes a prior determination of whether the agent's action satisfied the obligation, and none produces it. RAILS (highlighted) is the verification-native clearing layer that produces it; its band carries the seven primitives unfolded in detail by Figure~\ref{fig:rails-glance}, with the per-output RAILS Score as the Mesh's triage verifier and RAILS Report Cards as a model-scope source of reliability priors, and the evaluation-and-rating layer feeds it as verifier inputs and priors. The vertical annotations record the two organizing axes of the stack: trust requirement increases toward the top; evidence becomes available earlier toward the bottom.}
\label{fig:stack}
\end{figure}

Figure~\ref{fig:stack} places RAILS in the agentic stack. Above the clearing layer sit the protocols that act on the cleared decision: payment rails, and settlement and risk standards. Below sit the protocols that set up the obligation or produce the artifacts RAILS consumes: authorization and mandate, checkout and payment authorization, discovery and marketplace, the evaluation and rating layer, inter-agent communication, tool access, the model layer, and the substrate of compute, identity, and signatures. The vertical order is a dependency ordering, where each layer sits above what it drives or consumes; it is not a time order, because the money layers each act both before and after clearing. In time the sequence runs: the agent is authorized and an obligation is struck $\rightarrow$ checkout authorizes the charge and escrow locks the funds $\rightarrow$ the agent does the work $\rightarrow$ RAILS clears the result, releasing, refunding, or slashing $\rightarrow$ the settlement-risk standard applies that verdict $\rightarrow$ the funds are captured and moved $\rightarrow$ the outcome lands. Existing infrastructure has converged on the layers immediately above and immediately below; the clearing layer that connects them --- the layer that converts what happened into what should be settled --- has been missing.

Section 4 specifies the formal substrate on which the seven primitives are defined. Section 5 develops the admissibility-graded contribution that gives the protocol its soundness guarantee.

\section{Formal Model}

A protocol that cannot be implemented is a manifesto, not a specification. The seven primitives sketched in Section 3 become a specification only when their data shapes and the signatures of the functions that connect them are fixed. This section fixes them.

This section defines the obligation tuple, the execution trace and the Evidence Envelope, the admissibility-class function, the verifier signature, the mesh aggregator, the Clearing Decision, the Settlement Instruction, and the finality predicate. It does not prove that any of these structures has a particular property. The soundness statement that links emitted settlement to admissible evidence, the load-bearing claim of the paper, is the business of Section 5. Section 8's worked scenario instantiates the shapes specified here; Section 5 establishes the properties they must satisfy.

\subsection{Preliminaries}

Four substrate assumptions stand behind everything that follows: a public-key identity scheme that assigns each party, agent, verifier, and Clearing Engine instance a verification key bound to a stable identifier; a digital-signature scheme such that $\sigma_X$ over an object $o$ means the holder of $X$'s private key has endorsed $h(o)$; a collision-resistant hash function $h(\cdot)$ used uniformly for object commitments and foreign-key anchors; and a wall-clock time $t$ with monotonic semantics sufficient to enforce appeal windows.

These choices are deployment parameters. Any IETF-compliant stack for identity, signature, and hash satisfies the protocol's abstract requirements; the trust dependency this creates on the underlying public-key infrastructure is discussed in Section 12. The appendix records the specific algorithm choices used in the worked scenario.

\subsection{The Obligation Object}

The Obligation Object is the protocol's contract. It is a tuple:
\[
O = \langle \mathrm{id}_O, P, A, d, A^c, E^{req}, P_v, \varphi_O, P_s, P_f, \sigma_P, h_O \rangle
\]
where $\mathrm{id}_O$ is the obligation identifier; $P$ is the set of contracting parties, each carrying a public identity; $A$ is the set of agent identities authorized to act under $O$ (typically delegates from parties in $P$); $d$ is the task statement, expressed in natural language and optionally in formal predicates over a domain ontology; $A^c$ is the acceptance-criteria predicate over evidence the obligation will produce; and $E^{req}$ is the typed list of evidence items the obligation requires, each entry naming a class, a source, and a minimum admissibility floor where applicable.

$P_v$ is the verifier policy: a configuration that selects verifiers from a registry, fixes the mesh's composition, and specifies the rotation and quorum rules the aggregator will read. $\varphi_O$ is the obligation's admissibility floor, an element of $\Lambda$, the partial order on admissibility classes whose structure Section 5 develops. $P_s$ is the settlement policy: a rule mapping clearing-decision outcomes to settlement-instruction templates over the settlement systems the obligation permits. $P_f$ is the finality policy: the parameters of the finality predicate $\phi$, including a confidence threshold $c_{\min}$ and an appeal window $\tau_a$. $\sigma_P$ is the set of party signatures over $\bar{O}$, the obligation stripped of its hash field. $h_O = h(\bar{O})$ is the obligation hash: the value every downstream primitive uses to anchor itself to $O$.

\subsection{Execution Trace and Evidence Envelope}

The execution trace $T_O$ is the (in general, unobservable to the protocol) record of what the agent actually did under $O$: tool invocations, state changes, intermediate decisions. The protocol does not consume the trace directly. It consumes the Evidence Envelope, which is what the executing party and any third-party emitters chose to make admissible from the trace.
\[
E = \langle h_O, \tau, \langle e_1, \ldots, e_n \rangle, \Pi, \sigma_E \rangle
\]
Here $h_O$ anchors $E$ to its obligation; $\tau$ is the trace identifier, a deployment-stable handle that uniquely identifies the execution episode; each $e_i$ is an evidence item carrying a payload and metadata; $\Pi = \langle \pi_1, \ldots, \pi_n \rangle$ is the provenance chain assigning to each item the sequence of emitters and their signatures; $\sigma_E$ is the set of envelope-level signatures from the assembling parties.

Each item carries an admissibility class assigned by a function
\[
\mathrm{cls} : \text{Items} \to \Lambda
\]
where $\Lambda$ is a poset of admissibility classes. Section 5 develops $\Lambda$: its elements, the partial-order relation $\preceq$, and the join operation $\bigsqcup$ used to combine evidence-item classes into a basis class. For Section 4, only the signature of $\mathrm{cls}$ matters. Every evidence item lands in some class in $\Lambda$; Section 5 fixes what those classes are and how they compare.

\subsection{Verifier Signature}

A verifier $v_i$ is a function:
\[
v_i : (O, E) \to (s_i, c_i, B_i, \ell_i, r_i)
\]
where $s_i$ is the verdict, drawn from a per-deployment alphabet typically containing at minimum PASS, FAIL, ABSTAIN, and DISPUTED; $c_i \in [0, 1]$ is the verifier's confidence in $s_i$; $B_i \subseteq \{e_1, \ldots, e_n\}$ is the declared evidence basis, the subset of $E$ the verifier relied on to produce its verdict; $\ell_i$ is the verifier's loss estimate, defined when the verdict implies harm and undefined otherwise; and $r_i$ is the verifier's role tag: receipt verifier, constraint verifier, semantic judge, policy checker, human arbiter, or any extension class a deployment defines.

Two facts about the verifier contract matter for what follows. Each verifier carries a reliability prior $W_i$ drawn from its Clearing Passport, read separately from $v_i$'s output; the aggregator may weight verifier verdicts by this prior. The protocol assumes verifier integrity: $v_i$ correctly reports $B_i$. Verifiers that misreport their basis are an adversary class (basis-laundering), handled in Section 10.

\subsection{Mesh Aggregator}

The mesh aggregator combines the outputs of the $k$ verifiers selected by $P_v$:
\[
\Gamma : (\{v_i(O, E)\}_{i=1..k}, W, \Pi_v) \to (p_{\mathrm{perf}}, p_{\mathrm{pol}}, \hat{F}, c, B, \ell)
\]
$W = \langle W_1, \ldots, W_k \rangle$ is the verifier-reliability prior vector. $\Pi_v$ is the set of mesh-level parameters: rotation rules, quorum requirements, conflict-resolution procedures. $p_{\mathrm{perf}}$ and $p_{\mathrm{pol}}$ are the aggregate performance and policy verdicts; $\hat{F}$ is the fault-assignment record derived from $V_{\mathrm{out}}$ under the rules in $\Pi_v$; $c$ is the aggregate confidence; $B$ is the aggregate evidence basis derived from the surviving verifiers' bases; $\ell$ is the aggregate loss estimate.

The aggregate basis class is the join in $\Lambda$ of the classes of the items in $B$:
\[
\mathrm{cls}(B) = \bigsqcup_{e \in B} \mathrm{cls}(e)
\]
This is the class of the cross-verifier aggregate $B$, the union of the surviving verifiers' bases. The class of a single verifier's declared basis $B_i$ is instead the meet of its items, since $B_i$ names the evidence the verifier relied on and a verdict is only as admissible as the weakest evidence behind it; Section 5.1 develops the distinction.

Section 5 specifies the properties $\Gamma$ must satisfy (admissibility monotonicity and floor enforcement chief among them) and identifies admissibility-weighted majority as the reference aggregator. Here, the signature is the contract.

\subsection{Clearing Decision and Settlement Instruction}

The Clearing Decision is the canonical structured output of the mesh-and-aggregator pipeline:
\[
CD = \langle h_O, p_{\mathrm{perf}}, p_{\mathrm{pol}}, \hat{F}, \ell, B, \mathrm{cls}(B), V_{\mathrm{out}}, c, f, \sigma_\Gamma, h_{CD} \rangle
\]
$h_O$ anchors $CD$ to its obligation. $p_{\mathrm{perf}}$ and $p_{\mathrm{pol}}$ are the performance and policy verdicts: did the agent complete the task, and did the agent stay within the rules. $\hat{F}$ is the fault-assignment record; $\ell$ is the aggregate loss estimate; $B$ is the aggregate evidence basis; $\mathrm{cls}(B)$ is its class in $\Lambda$; $V_{\mathrm{out}}$ is the record of individual verifier outputs, preserved for audit. $c$ is the aggregated confidence; $f$ is the finality status, initially PROVISIONAL; $\sigma_\Gamma$ is the Clearing Engine's signature, with optional co-signatures from attesting verifiers; $h_{CD} = h(\overline{CD})$ is the clearing-decision hash.

A $CD$ is \textbf{non-DISPUTED} iff neither $p_{\mathrm{perf}}$ nor $p_{\mathrm{pol}}$ equals DISPUTED; otherwise it is DISPUTED. Settlement Instruction emission (Section 5.4) is gated on non-DISPUTED $CD$s.

The Settlement Instruction translates $CD$ into rail-executable actions:
\[
S = \langle h_{CD}, A_{\mathrm{fee}}, A_{\mathrm{prin}}, A_{\mathrm{coll}}, A_{\mathrm{claim}}, A_{\mathrm{pen}}, A_{\mathrm{rep}}, R, \rho, \sigma_\Gamma, h_S \rangle
\]
$h_{CD}$ anchors $S$ to its clearing decision. The six action fields (fee, principal, collateral, claim, penalty, reputation) are populated by $P_s$ applied to $CD$. $R$ is the execution-rail identifier; $\rho$ is the receipt requirement specifying whether and how the rail acknowledges settlement back into the RAILS Envelope. $\sigma_\Gamma$ is the Clearing Engine's signature, and $h_S = h(\bar{S})$ is the instruction's hash.

\subsection{Finality}

Finality is a predicate over a Clearing Decision, the current time, and the post-emission environment $\varepsilon$ (which records appeals filed and conflicts pending):
\[
\phi(CD, t, \varepsilon) = (\mathrm{cls}(B) \succeq \varphi_O) \wedge (c \geq c_{\min}) \wedge \mathrm{NoUnresolvedConflict}(V_{\mathrm{out}}, \varepsilon) \wedge (t \geq t_{\mathrm{emit}} + \tau_a)
\]
$c_{\min}$ and $\tau_a$ come from $P_f$. When $\phi$ fires, $f$ transitions from PROVISIONAL to FINAL. Until then, the Settlement Instruction's effects are reversible: settlement systems that have executed PROVISIONAL settlements must support reversal until $f = \mathrm{FINAL}$.

The four conjuncts correspond to four ways finality can be blocked: the basis was too weak (admissibility), the verdict was too uncertain (confidence), the verifiers disagreed substantively (conflict), or the appeal window has not yet elapsed (time). The conjunction is strict; satisfying three of four is not sufficient. Section 6 develops the lifecycle in which $f$ transitions occur (Figure~\ref{fig:lifecycle}).

\medskip

A notation table summarizing the symbols introduced in Section 4 appears in the appendix. Figure~\ref{fig:objects} shows the four core data objects (Obligation Object, Evidence Envelope, Clearing Decision, Settlement Instruction) and the hash-anchor and signature-chain relationships among them.

\begin{figure}
\centering
\includegraphics[width=\linewidth,keepaspectratio]{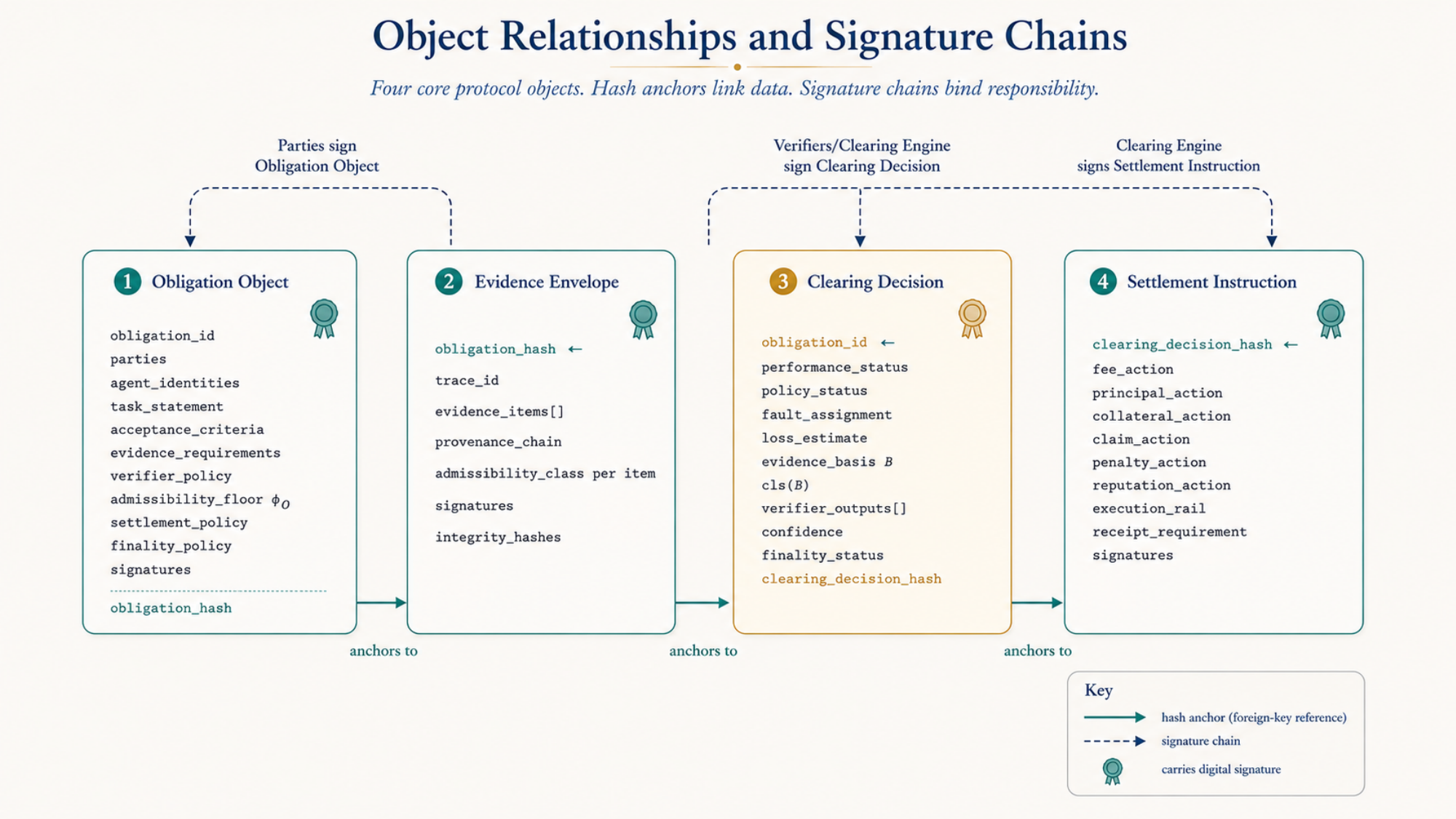}
\caption{Hash-anchor and signature-chain relationships among the four central RAILS objects. Each downstream object anchors to its predecessor by hash reference (Evidence Envelope to Obligation Object; Clearing Decision to Obligation Object; Settlement Instruction to Clearing Decision); canonical field names and types appear in Appendix A.3. Signatures chain across the objects: parties sign the Obligation Object; verifiers and the Clearing Engine co-sign the Clearing Decision; the Clearing Engine alone signs the Settlement Instruction. The Clearing Decision (highlighted) is the canonical protocol output that downstream rails commit to.}
\label{fig:objects}
\end{figure}

With these signatures in place, the soundness statement RAILS imposes on $\Gamma$ (that any emitted Settlement Instruction's basis class meets the obligation's admissibility floor) becomes a clean mathematical condition:
\[
\mathrm{Emit}(S) \implies \mathrm{cls}(B) \succeq \varphi_O
\]
Section 5 establishes this property.

\section{The Admissibility-Graded Verification Mesh}

Section 4 closed with a forward pointer: that the soundness statement RAILS imposes on its mesh aggregator ($\mathrm{Emit}(S) \implies \mathrm{cls}(B) \succeq \varphi_O$) follows from the contracts this section specifies. This section establishes it.

The argument has four moves. Section 5.1 develops $\Lambda$ as a partial order (not a scalar ranking) and explains why incomparability is irreducible. Section 5.2 fixes the contract every verifier must satisfy: a declared evidence basis, honestly reported. Section 5.3 specifies $\Gamma$'s two required properties: admissibility monotonicity and floor enforcement. Section 5.4 states the soundness statement and sketches the one-line argument that establishes it. The three subsections that follow trace the consequences.

The soundness statement is falsifiable against the specification.

\subsection{The admissibility partial order \texorpdfstring{$\Lambda$}{Lambda}}

Admissibility classes form a partial order, not a scalar ranking. $\Lambda$ contains six canonical classes: SELF (unverified self-report by the acting agent), SIGN (cryptographically signed by the acting agent: non-repudiation, not truth), WIT (signed by a third-party witness present to the action, human or institutional), REC (signed receipt from a non-interested external system), ATT (attestation from a trusted execution environment or similarly hardened runner), and PROOF (cryptographic proof of correctness, such as zk-SNARK or formal verification certificate). The covering relations are:
\[
\mathrm{SELF} \preceq \mathrm{SIGN} \preceq \{\mathrm{WIT}, \mathrm{REC}\} \preceq \mathrm{ATT} \preceq \mathrm{PROOF}
\]
WIT and REC are deliberately incomparable. A third-party witness signature is not strictly stronger or weaker than a non-interested external receipt: the two verify different things. A witness attests to having observed an event; a receipt records that an external system processed a transaction. A scalar order would force a comparison the underlying evidence does not support.

The class of an evidence item with a multi-hop provenance chain is the class of its weakest link: an attested artifact rebroadcast through a self-reporting channel has the class of the rebroadcast, not the attestation. This makes provenance verification at envelope ingest the load-bearing audit step, and the precondition for soundness that Section 5.7 names.

Deployments may extend $\Lambda$ for domain-specific evidence (adding, for example, a notarization class above WIT or a regulator-attested receipt above ATT), provided the result remains a poset with top PROOF and bottom SELF. The join used in Section 4 computes the class of the aggregate basis $B$ across the surviving verifiers; the class of a single verifier's declared basis is the meet of the items it relied on, as the next paragraph develops.

Three combinations appear in the protocol, and they are not the same operation. A multi-hop provenance chain takes the meet: an evidence item is only as admissible as the weakest link it passed through, so an attested artifact rebroadcast through a self-reporting channel has the class of the rebroadcast. A single verifier's declared basis also takes the meet: because $B_i$ is the set of items the verifier relied on, a verdict that leaned on a sub-floor item is only as admissible as that item, and the basis class is the weakest item in $B_i$. The aggregate basis $B$, taken across the verifiers that survive floor enforcement, takes the join: once every surviving verifier has individually cleared the floor, the strongest of their bases summarizes the result. The meet protects the floor; the join only reports it. Combination never promotes a basis: a verifier that relied on a WIT item and a REC item has basis class SIGN, their meet, not ATT, so no set of individually sub-floor items can clear a floor by being consulted together.

\begin{figure}
\centering
\includegraphics[width=\linewidth,height=0.72\textheight,keepaspectratio]{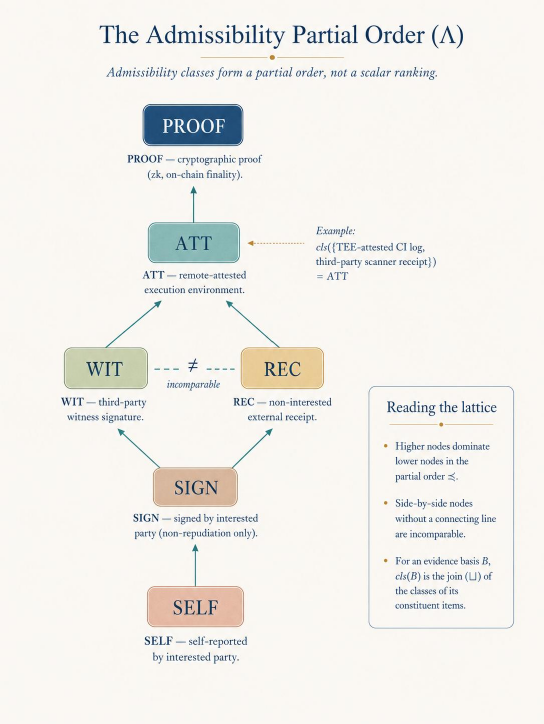}
\caption{The admissibility partial order $\Lambda$. Six classes form a partially ordered set, not a total order: SELF (self-report by the acting agent), SIGN (cryptographic signature by the acting agent), WIT (third-party witness signature), REC (non-interested external receipt), ATT (attestation from a trusted execution environment), and PROOF (cryptographic proof of correctness). WIT and REC are incomparable; the protocol records that incomparability rather than collapsing the two into a scalar, because witness signatures and third-party receipts rest on structurally different trust assumptions. The class of an aggregate basis $B$ is the join in $\Lambda$ of the classes of its constituent items (the example on the right shows $\mathrm{cls}(\{\text{TEE-attested CI log}, \text{third-party scanner receipt}\}) = \mathrm{ATT}$). Deployments may extend $\Lambda$ for domain-specific evidence; the protocol requires only that the result remain a poset with top PROOF and bottom SELF.}
\label{fig:lambda}
\end{figure}

Figure~\ref{fig:lambda} shows $\Lambda$ as a Hasse diagram.

\subsection{The declared basis as a load-bearing contract}

Recall the verifier signature from Section 4:
\[
v_i : (O, E) \to (s_i, c_i, B_i, \ell_i, r_i)
\]
The component that matters most for what follows is $B_i$, the declared evidence basis. It is the verifier's commitment about which evidence items support its verdict. Without it, the aggregator has no way to compute $\mathrm{cls}(B)$, and the soundness statement becomes unenforceable.

The protocol assumes verifier integrity: $v_i$ correctly reports $B_i$. This is a contract, not a guess about the world. Every verifier admitted to a RAILS mesh accepts the obligation to declare its basis honestly: to include in $B_i$ every evidence item it relied on, and to exclude every item it did not. A verifier that includes evidence it ignored, or omits evidence it consulted, has broken the contract.

Verifiers that misreport are an adversary class (basis-laundering), and Section 10 develops the defenses. The contract is the price of admission to the protocol, and it is the precondition for everything that follows.

\subsection{The Mesh Aggregator \texorpdfstring{$\Gamma$}{Gamma} and its required properties}

Section 4 fixed $\Gamma$'s signature. Section 5.3 fixes what any compliant $\Gamma$ must guarantee. Two properties are required.

\textbf{Admissibility monotonicity.} Upgrading any $\mathrm{cls}(e)$ under $\preceq$ cannot reduce $c$ for the same verdict $p$. Aggregators that punish stronger evidence are pathological: they discourage submission of better-attested artifacts and create perverse incentives for parties to withhold corroborating evidence. Monotonicity is the structural fix. A verifier that submits the same verdict on a stronger basis must, under any compliant $\Gamma$, receive a weight at least as large. The protocol does not specify the functional form of the weight, only that it is monotone in $\mathrm{cls}$ under $\preceq$.

\textbf{Floor enforcement.} $\Gamma$ may not return a non-DISPUTED verdict with $c$ above threshold if $\mathrm{cls}(B)$ is below $\varphi_O$. Inadmissible mesh outputs are downgraded to UNVERIFIABLE or DISPUTED, and the Settlement Instruction emission path is gated on a non-DISPUTED verdict. This is not a tunable parameter or a deployment preference; it is the operational basis for the soundness guarantee. A $\Gamma$ that emits a non-DISPUTED verdict on a sub-floor basis is, by definition, not a RAILS-compliant aggregator.

The protocol's reference aggregator is admissibility-weighted majority. Each verifier's weight is:
\[
\mathrm{weight}_i = W_i \cdot \mathbb{1}[\mathrm{cls}(B_i) \succeq \varphi_O]
\]
where $\mathbb{1}[\cdot]$ is the indicator function. Verifiers whose declared basis falls below the floor receive weight zero: their verdict does not count toward the aggregate. Here $\mathrm{cls}(B_i)$ is the meet of the items in $B_i$ (Section 5.1), so $\mathrm{cls}(B_i) \succeq \varphi_O$ holds exactly when every item the verifier relied on meets the floor; a verifier that leaned on even one sub-floor item receives weight zero. Among the verifiers whose basis meets or exceeds the floor, weights are scaled by the reliability prior $W_i$ drawn from each verifier's Clearing Passport. The aggregate verdicts $(p_{\mathrm{perf}}, p_{\mathrm{pol}})$ are the majority verdicts under these weights; the aggregate confidence $c$ is a function of the weight-mass behind the majority (the precise functional form is a deployment parameter, constrained by monotonicity); the aggregate basis $B$ is the union of the bases of the surviving verifiers (the verifiers whose weight is nonzero).

The protocol fixes how priors are used but not how they are computed. Populating them well is non-trivial and is where calibrated scoring systems, per-model, per-task, grounded in measured failure rates, do their work. The protocol is open at this seam by design; the calibration that fills it is a deployment's own contribution.

This is one valid aggregator. Others are possible: an admissibility-weighted Bayesian aggregator, a unanimity-required aggregator for the highest-exposure obligations, a learning-augmented aggregator that updates $W_i$ online. The protocol fixes the properties; the reference aggregator is normative for the default deployment but not for all.

\begin{figure}
\centering
\includegraphics[width=\linewidth,keepaspectratio]{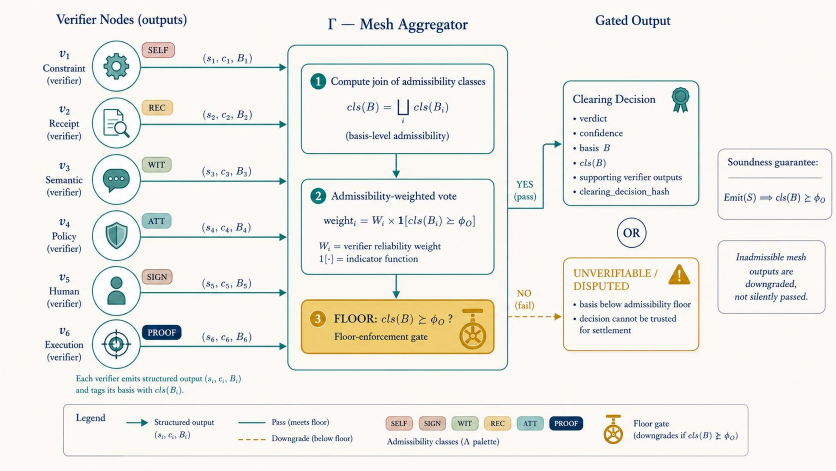}
\caption{The admissibility-weighted Mesh Aggregator $\Gamma$. Each verifier $v_i$ emits a structured output $(s_i, c_i, B_i)$: a verdict, a confidence, and a declared evidence basis whose admissibility class $\mathrm{cls}(B_i)$ sits in $\Lambda$. $\Gamma$ first computes the basis-level admissibility $\mathrm{cls}(B)$ as the join of the surviving verifiers' classes, then runs an admissibility-weighted vote in which verifiers below the floor contribute zero weight. The floor-enforcement gate (gold) is what enforces the soundness guarantee $\mathrm{Emit}(S) \implies \mathrm{cls}(B) \succeq \varphi_O$: when $\mathrm{cls}(B) \succeq \varphi_O$ holds, the aggregate becomes a Clearing Decision; otherwise the output is downgraded to UNVERIFIABLE/DISPUTED. Inadmissible mesh outputs are downgraded rather than silently passed; the gate is the operational locus of the protocol's single load-bearing property.}
\label{fig:gamma}
\end{figure}

Figure~\ref{fig:gamma} shows the aggregator's data flow, with the floor-enforcement gate rendered as the gold valve at the bottom of $\Gamma$.

\subsection{The soundness statement}

With $\Lambda$ developed (Section 5.1), the declared-basis contract fixed (Section 5.2), and $\Gamma$'s required properties specified (Section 5.3), the protocol's central soundness property follows directly.

\begin{quote}
\textbf{Soundness.} For any obligation $O$ with admissibility floor $\varphi_O$, any Settlement Instruction $S$ the Clearing Engine emits under $O$ satisfies:
\[
\mathrm{Emit}(S) \implies \mathrm{cls}(B) \succeq \varphi_O.
\]
\end{quote}

\textbf{Proof.} Settlement Instructions are emitted only on non-DISPUTED Clearing Decisions (Section 4.6). By $\Gamma$'s floor-enforcement clause, neither $p_{\mathrm{perf}}$ nor $p_{\mathrm{pol}}$ attains a non-DISPUTED status above threshold unless $\mathrm{cls}(B) \succeq \varphi_O$. The verifier-integrity assumption guarantees that each $B_i$, and therefore the aggregate $B$ derived from the surviving verifiers, is reported honestly. Therefore $\mathrm{Emit}(S) \implies \mathrm{cls}(B) \succeq \varphi_O$. The result follows from the floor-enforcement clause and the declared-basis contract. The proof is immediate from the definitions: floor enforcement is specified to exclude sub-floor bases, so soundness records that the gate does what it is defined to do rather than establishing a deep result. Its force is in the precision of the claim, not the length of its proof.

The statement is falsifiable against the specification: a reviewer can construct an aggregator $\Gamma'$ that satisfies $\Gamma$'s signature but violates floor enforcement, forcing the specification to either tighten the property or acknowledge the gap. That falsifiability is what separates a specified protocol from a manifesto.

This property also marks the line between RAILS and the verification mechanisms nearest to it. Three families of prior approach gate settlement on a check of some kind. Predicate-based gating releases funds when an execution proof satisfies a fixed boolean condition, so the output is a pass with no notion of how strong the evidence behind it was. Score-based adjudication aggregates verifier or validator opinions into a numeric quality rating, so the output is a number with no specification to falsify it against. Challenge-game correctness makes honesty the cheaper strategy through staking and slashing, so the guarantee is an equilibrium that holds under rationality assumptions rather than a property of the emitted artifact. RAILS differs in the object of the guarantee. The soundness statement is neither a pass, nor a score, nor an equilibrium. The score it is not is a verdict emitted as the answer; the per-output RAILS Score of Section 5.6 is by contrast a verifier input to this determination, not a verdict in its own right. It is a falsifiable claim about every emitted Settlement Instruction, refutable by constructing an aggregator that satisfies the signature and violates floor enforcement. Graded admissibility supplies the structure these three families lack, an explicit floor below which a material settlement cannot be supported.

Soundness applies at every emission, PROVISIONAL or FINAL. FINAL emission additionally requires the finality predicate $\phi$ to fire (the four conjuncts of Section 4.7), of which admissibility-floor satisfaction is one.

\subsection{What soundness gives and does not give}

Soundness gives a clean line between admissible and inadmissible settlement. No financially material Settlement Instruction is emitted on evidence below the obligation's declared floor. The protocol's audit trail is admissible by construction.

Soundness does not give verifier correctness. A verifier can honestly report its basis and still produce a wrong verdict on that basis. Calibration is a property of the verifier, not the protocol; it is addressed empirically across transactions by Clearing Passports, not guaranteed per-decision.

Soundness does not give ground truth. Whether the agent actually satisfied the obligation, in the world, is outside the protocol's reach. RAILS adjudicates on evidence; what happened is what the evidence is evidence about.

Soundness does not give attestor honesty. A corrupted TEE attestor can emit bogus ATT-class evidence; the trust root sits outside the protocol. Soundness also does not give validity of $\varphi_O$ itself: whether the parties set the right floor is a governance question, not a protocol question.

The analogy is cryptographic soundness in the sense familiar from zero-knowledge proofs. A sound ZK system cannot be convinced of a false statement via the protocol; false beliefs outside the protocol remain possible. RAILS soundness is the same shape: it bounds what the protocol can be made to emit, not what is true in the world.

\subsection{A catalog of verifier classes}

The verifier signature is uniform (every verifier produces the same five-tuple), but verifier types differ in what they look at and how they typically fail. The protocol does not mandate any specific set of verifier classes; obligations select whatever combination of verifiers their verifier policy $P_v$ specifies. The five roles below recur across deployments and are the reference classes the worked scenario in Section 8 instantiates. Deployments may add others: Ranking Inference (RI), the token-rank scoring primitive of \cite{bogdan2026mandelbrot}, is a candidate first-pass extension that adds CPU-only triage scoring over token-level outputs upstream of more expensive verifiers.

\begin{table}[h]
\centering
\begin{tabular}{p{1.8cm}>{\raggedright\arraybackslash}p{3.6cm}>{\raggedright\arraybackslash}p{3.2cm}p{4.2cm}}
\toprule
\textbf{Role tag} & \textbf{Verifies} & \textbf{Typical basis class} & \textbf{Characteristic failure} \\
\midrule
constraint & Deterministic predicates over evidence (scope, dependency policy, format) & ATT or higher (predicate over signed artifacts) & False negatives when constraints are under-specified \\
receipt & Receipts from non-interested external systems (CI runners, scanners, ledgers) & REC or ATT (depending on receipt origin) & Receipt forgery; receipt staleness \\
semantic & Open-ended performance evaluation, typically LLM-based & Often heterogeneous; SELF-class self-reports lower the basis & Bias (position, verbosity, self-enhancement); prompt injection; basis-laundering when judges include SELF items \\
policy & Compliance with policies (authority, content, jurisdiction) & Mandate evidence (WIT or ATT) & Policy gaps; post-hoc policy changes \\
human & Human arbitration, typically in escalation or appeal & Whatever the arbiter is shown; signed approval (WIT) & Cost; latency; reviewer inconsistency \\
\bottomrule
\end{tabular}
\end{table}

The table is descriptive, not normative. Any verifier conforming to the signature may participate in a RAILS Mesh, and the obligation's verifier policy $P_v$ determines which classes (and how many of each) run for any given clearing event. The aggregator does not care what role tag a verifier carries; it cares only about the verifier's reliability prior $W_i$ and the admissibility class of its declared basis.

The catalog above shares a structural blind spot. Constraint, receipt, semantic, policy, and human verifiers each assess the output in front of them; none assesses the behavioral reliability of the model that produced it, whether a model of this type, on this class of task, tends toward sycophantic agreement, instability under input perturbation, or weak grounding. This is a distinct axis of evidence, orthogonal to whether a given output meets the acceptance criteria, and a clearing deployment operating over agent outputs at scale requires a verifier class that supplies it. The measurement substrate for such a class is already established. The real-time Ranking Inference verification primitive of \cite{bogdan2026mandelbrot} computes output-scope reliability signals at CPU-only cost, and the Machine Psychometrics behavioral profiling of \cite{bogdan2026mindprints} computes model-scope susceptibility signals such as sycophancy and perturbation instability; together they specify how the reliability of a generating model is measured and calibrated rather than assumed. A verifier built on those frameworks enters the Mesh under the same signature as any other, returning a verdict, a confidence, and a declared basis, yet it contributes evidence no other class in this catalog can produce.

We call this per-output composite the RAILS Score: a reliability estimate computed on every output, composed of multiple sub-scores such as the real-time Ranking Inference verification primitive of \cite{bogdan2026mandelbrot} and the Machine Psychometrics behavioral profiling of \cite{bogdan2026mindprints}. It enters the Mesh under the standard verifier signature, serves as the always-on triage verifier of the default tier (Section 7.2), and is the first primitive shipped in the deployment ladder of Section 13.

\subsection{Threat-model corollary}

The soundness statement has three preconditions: that $\Lambda$ is correctly assigned to evidence, that verifiers honestly declare their bases, and that the obligation binds the floor the parties intended. Each precondition is a target. To produce an unsound settlement under RAILS, an attacker must succeed at at least one of:

\begin{itemize}
\tightlist
\item \textbf{FORGE-UP}: claim a higher $\mathrm{cls}(e)$ than the provenance chain supports. Defeated by provenance-chain verification at envelope ingest.
\item \textbf{LAUNDER-BASIS}: collude with a verifier to misreport $B_i$. Defeated by verifier rotation, basis-disclosure audits, and Clearing Passport drift detection.
\item \textbf{DOWNGRADE-FLOOR}: cause obligation-binding to settle on a weaker $\varphi_O$ than the parties intended. Defeated by obligation-binding signature requirements and template review.
\end{itemize}

Section 10 develops each family in admissibility space rather than as a list of disconnected attacks, and maps previously-reported attacks from the literature to instances of these three. The worked scenario in Section 8 shows the floor-enforcement clause acting on a SELF-tainted verifier basis without any attack present: an honest verifier's honest disclosure of a weak basis, correctly downgraded by the aggregator. That non-adversarial case is the easy one; Section 10 is the hard one.

\section{Lifecycle State Machine}

The data objects of Section 4 do not exist in a vacuum. Each comes into being at a specific stage of a process; each lives through specific transitions; each may reach specific terminations. The soundness statement of Section 5 applies at every emission of a Settlement Instruction, but emissions occur at a specific point in a longer process, and the protocol's correctness depends on the structure of that process.

\begin{figure}
\centering
\includegraphics[width=\linewidth,keepaspectratio]{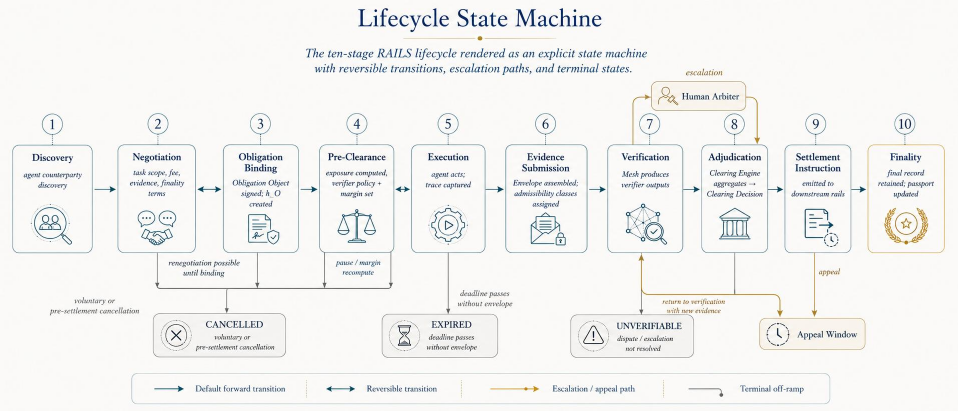}
\caption{The RAILS lifecycle as a state machine. Ten primary states run from Discovery through Negotiation, Obligation Binding, Pre-Clearance, Execution, Evidence Submission, Verification, Adjudication, and Settlement Instruction to Finality (gold-accented). Two reversibility boundaries permit return transitions before the obligation binds (Negotiation $\leftrightarrow$ Obligation Binding) and before evidence is submitted (Pre-Clearance $\leftrightarrow$ Execution); two escalation loops route Verification into a Human Arbiter side-node and Adjudication into an Appeal Window. Three terminal off-ramps below the main flow (CANCELLED, EXPIRED, UNVERIFIABLE) record the lifecycles that do not produce a clearing decision; everything else converges on Finality, the only state in which a settled obligation becomes binding.}
\label{fig:lifecycle}
\end{figure}

This section fixes the structure: ten stages from initial counterparty discovery to final settlement, two reversibility boundaries, two escalation loops, and three terminal exits that are not Finality. Figure~\ref{fig:lifecycle} renders the lifecycle as a state machine; the prose below traces it in order.

\subsection{The ten stages}

\begin{enumerate}
\tightlist
\item \textbf{Discovery.} A requesting party finds candidate counterparties: agents (or other parties) able and willing to undertake the obligation. Discovery is governed by the marketplace or directory the parties operate within; what enters the lifecycle from this stage is a candidate set of agents, not yet bound.
\item \textbf{Negotiation.} Parties propose, counter, and converge on the task statement, the scope and prohibitions, the acceptance criteria, the evidence requirements, the fee, the admissibility floor $\varphi_O$, and the settlement and finality policies. Nothing is signed; nothing is binding.
\item \textbf{Obligation Binding.} Parties sign the Obligation Object; the obligation hash $h_O$ is computed; the binding event is published to the parties' Clearing Passports. From this stage forward, the data object $O$ is canonical and immutable.
\item \textbf{Pre-Clearance.} Exposure is computed for $O$ (see Section 7); the verifier policy $P_v$ is configured against the verifier registry; margin requirements are posted by the responsible party. The Mesh is selected but not yet running.
\item \textbf{Execution.} The authorized agent acts under $O$. The execution trace $T_O$ accumulates (tool calls, state changes, intermediate decisions), most of which the protocol does not consume directly.
\item \textbf{Evidence Submission.} The executing party and any third-party emitters assemble the Evidence Envelope $E$ from $T_O$, sign it, and submit it to the Mesh. Each evidence item is tagged with its admissibility class at submission.
\item \textbf{Verification.} The Verification Mesh runs the verifiers selected by $P_v$. Each verifier returns its tuple $(s_i, c_i, B_i, \ell_i, r_i)$. Verifiers run independently and may operate concurrently.
\item \textbf{Adjudication.} The Clearing Engine $\Gamma$ aggregates the verifier outputs under the obligation's admissibility floor, computes $\mathrm{cls}(B)$, applies the floor-enforcement clause, and emits a Clearing Decision $CD$ whose performance and policy verdicts $(p_{\mathrm{perf}}, p_{\mathrm{pol}})$ each lie in \{PASS, FAIL, ABSTAIN, DISPUTED\}.
\item \textbf{Settlement Instruction.} If $CD$ is non-DISPUTED, the obligation's settlement policy $P_s$ is applied to produce a Settlement Instruction $S$, which is signed by the Clearing Engine and dispatched to the execution rail $R$. The instruction enters the rail in PROVISIONAL state.
\item \textbf{Finality.} The finality predicate $\phi$ from Section 4.7 is evaluated against the post-emission environment as time elapses and any appeals are filed or resolved. When $\phi$ fires, $f$ transitions from PROVISIONAL to FINAL, the Clearing Passports of the participants are updated, and the lifecycle terminates.
\end{enumerate}

\subsection{Reversibility}

The default direction of the lifecycle is forward. Two transitions are reversible, and these are the only two.

\begin{enumerate}
\tightlist
\item \textbf{Negotiation $\leftrightarrow$ Obligation Binding.} Parties may renegotiate freely until the Obligation Object is signed. Once signed, the obligation is immutable; further changes require a fresh binding event with a new $h_O$.
\item \textbf{Pre-Clearance $\leftrightarrow$ Execution.} Execution may be paused for margin recomputation if exposure shifts mid-task: for example, when a downstream policy change alters the obligation's admissibility floor mid-execution. The protocol permits a pause-and-recompute cycle without invalidating the trace, provided the recomputed margin is posted before execution resumes.
\end{enumerate}

All other transitions are forward-only. Settlement is reversible during the PROVISIONAL period (settlement systems must support reversal until $f = \mathrm{FINAL}$, per Section 4.7), but the lifecycle itself does not move backward from Settlement Instruction or Finality. Once finality is reached, the only further state change is the passport update, and that is part of finality, not a separate stage.

\subsection{Escalation paths and terminal exits}

Two escalation side-loops handle non-routine adjudication.

\begin{enumerate}
\tightlist
\item \textbf{Human Arbiter loop.} Verification may escalate to a human arbiter when the mesh cannot reach the confidence threshold under any conflict-resolution rule in $P_v$. The arbiter is itself a verifier in role-tag terms ($r = \text{human}$), and its output re-enters Adjudication as one more verifier vote, subject to the same floor-enforcement clause as any other verifier.
\item \textbf{Appeal Window loop.} A party may file an appeal against a PROVISIONAL Settlement Instruction within the appeal window $\tau_a$ from $P_f$. An appeal returns the obligation to Verification with whatever new evidence the appeal introduces, and a new Clearing Decision must be emitted before finality can be reached.
\end{enumerate}

Three terminal exits are not Finality. \textbf{CANCELLED}: parties agree to abort before settlement; collateral is released, no passport delta beyond the cancellation event. \textbf{EXPIRED}: the execution deadline passes without an Evidence Envelope; the obligation closes as default-by-non-performance. \textbf{UNVERIFIABLE}: Adjudication produces a DISPUTED $CD$ that no escalation resolves; the obligation closes without a clearing. These are not failure modes of the protocol; they are the protocol's recognition that not every obligation is clearable.

\medskip

The lifecycle is what gives the formal model its operational meaning. The data objects of Section 4 and the soundness statement of Section 5 acquire their place in a process with explicit reversibility boundaries, explicit escalation paths, and explicit terminal exits. Section 7 develops the exposure-driven policy that determines, at Pre-Clearance, which verifiers run and what margin is posted.

\section{Exposure-Driven Verification Intensity}

\begin{figure}
\centering
\includegraphics[width=\linewidth,keepaspectratio]{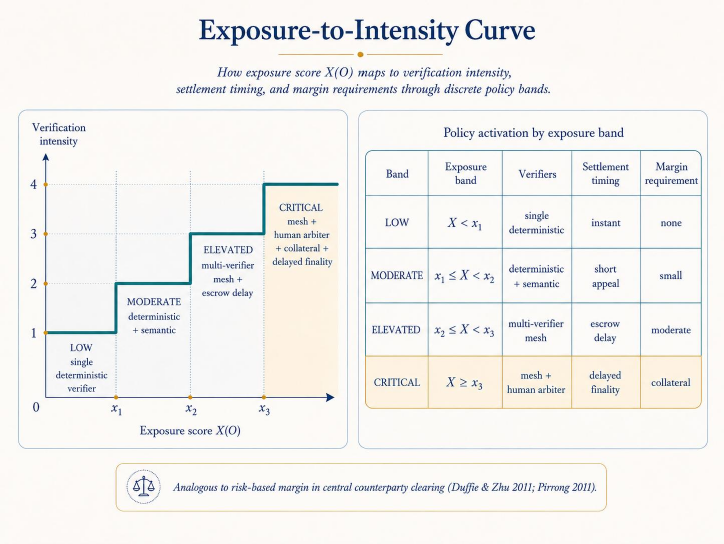}
\caption{The exposure-to-intensity policy. The protocol computes an exposure score $X(O)$ at Pre-Clearance from six standardized inputs (authority risk, value at risk, intent ambiguity, irreversibility, agent history, domain sensitivity). The policy $K = g(X, P_v, D)$ then maps $X$ to a verification regime through four discrete bands (LOW, MODERATE, ELEVATED, CRITICAL): the left panel renders $K$ as a stepped staircase climbing from a single deterministic verifier through multi-verifier mesh aggregation to mesh plus human arbiter; the right panel summarizes which verifiers, settlement timing, and margin requirement each band activates. The structure is the protocol's expressed tradeoff between verification cost and loss exposure, calibrated together against a single scalar; it is the agentic analogue of risk-based margin in central counterparty clearing~\cite{duffie2011ccp, pirrong2011clearing}.}
\label{fig:exposure}
\end{figure}

At Pre-Clearance (Section 6, stage 4), three decisions are made simultaneously: which verifiers will run, how long the Settlement Instruction will be held in PROVISIONAL, and what margin the responsible party must post. This section specifies the function that makes those three decisions. They are not independent. They are calibrated together against a single scalar that captures how much risk the obligation carries. That scalar is the exposure score $X(O)$; the policy $K$ it determines is a step function in $X$, not a continuous one; and the structure of the step function is the protocol's expressed tradeoff between verification cost and loss exposure. Figure~\ref{fig:exposure} shows the canonical staircase and the policy table that travels with it.

\subsection{The exposure score \texorpdfstring{$X(O)$}{X(O)}}

$X(O)$ is a scalar computed at Pre-Clearance from six standardized inputs. The protocol requires each be considered and requires $X$ to be monotone in each.

\begin{itemize}
\tightlist
\item \textbf{Authority risk}: the scope of what the agent is authorized to do under $O$. A read-only audit obligation carries less authority risk than an obligation that authorizes the agent to merge code to main, sign payments, or execute trades.
\item \textbf{Value at risk}: the principal exposed by the obligation: the fee, the claim ceiling, the collateral the responsible party posts against default.
\item \textbf{Intent ambiguity}: the clarity of the task statement and acceptance criteria. A sharply specified obligation with formal predicates over a domain ontology carries less ambiguity risk than a natural-language obligation with judgment-call acceptance criteria.
\item \textbf{Irreversibility}: whether the agent's actions can be unwound. Operations on append-only ledgers, public communications, and external counterparties are less reversible than file edits in a sandboxed branch.
\item \textbf{Agent history}: the Clearing Passport priors for this agent and its tool chain: prior dispute rate, prior basis-drift events, prior FORGE-UP attempts (Section 5.7).
\item \textbf{Domain sensitivity}: jurisdictional, regulatory, or reputational sensitivity. Medical, legal, financial, and safety-critical domains raise $X(O)$ by virtue of where the obligation lives, not by what the obligation asks.
\end{itemize}

The protocol fixes the inputs and requires monotonicity in each. The functional form is a deployment parameter. The reference deployment uses a normalized weighted sum; learned aggregators that satisfy the monotonicity constraint are admissible.

\subsection{Stepped policy bands}

The mapping from $X(O)$ to policy is a step function:
\[
K = g(X, P_v, D)
\]
where $P_v$ is the obligation's verifier policy (Section 4.2) and $D$ is the deployment configuration: the threshold values $x_1, x_2, x_3$ and the band-specific defaults. $K$ selects three coupled outputs:

\begin{enumerate}
\tightlist
\item The \textbf{verifier set}: which verifier classes from Section 5.6 to instantiate, how many of each, the quorum the aggregator will require.
\item The \textbf{settlement timing}: whether the Settlement Instruction executes instantly, holds for a short appeal window, escrows until external receipts confirm, or holds in PROVISIONAL for an extended deferred-finality interval.
\item The \textbf{margin requirement}: the collateral the responsible party must post against the obligation, ranging from none through fractions of principal to full collateralization.
\end{enumerate}

The four bands, as Figure~\ref{fig:exposure}'s right panel renders them:

\begin{itemize}
\tightlist
\item \textbf{LOW} ($X < x_1$): single deterministic verifier, instant settlement, no margin. The typical case for cheap, scoped, low-stakes obligations against high-history agents.
\item \textbf{MODERATE} ($x_1 \leq X < x_2$): deterministic plus semantic verifier, short appeal window, small margin. Most general-purpose obligations sit here.
\item \textbf{ELEVATED} ($x_2 \leq X < x_3$): multi-verifier mesh, escrow-delay settlement, moderate margin. Tasks with material principal, irreversibility, or ambiguity push obligations into this band.
\item \textbf{CRITICAL} ($X \geq x_3$): mesh plus human arbiter, delayed-finality settlement, full collateral. Reserved for high-authority or high-value obligations where the cost of being wrong dominates the cost of being slow.
\end{itemize}

A timing constraint underlies the entire staircase. Agentic commerce runs at machine speed: agents transact in the time of a model call, not the time of a human decision, and the bulk of obligations are low-exposure events that must clear inline or not clear at all. A verification layer that cannot return a verdict in real time on the common case cannot sit in the settlement path. The heavier verifier classes the staircase escalates to carry latency and cost that disqualify them as the always-on inline layer: a semantic LLM-judge runs an expensive model call and is the gameable class the protocol already guards against, an attestation-dependent receipt verifier waits on a trusted-execution or third-party round-trip, and a human arbiter resolves in minutes or hours. Each is an escalation-tier resource, justified only when exposure warrants its delay, which is why the staircase introduces them progressively rather than by default. The default tier, where most machine-speed volume settles, therefore requires a verifier that runs on every output in real time at minimal cost, computes an inline risk estimate, and decides which obligations escalate into the slower and more expensive tiers. Real-time triage is in this sense not an optimization of the Mesh but a precondition for clearing to operate at the speed agentic commerce runs.

Why stepped, not continuous: verifier instantiation is discrete; settlement-timing rules come from a finite library of policy templates; margin schedules are negotiated in tiers. A continuous $K$ would imply discriminations the underlying mechanisms cannot support. The thresholds $x_1, x_2, x_3$ are deployment-tunable; the four band names and the structure of the gradient are protocol-canonical.

\subsection{The CCP analogy}

RAILS's exposure-to-policy curve is the agentic analogue of risk-based margining in central counterparty clearing. A CCP requires posted collateral proportional to the risk of a member's position, with margin multipliers triggering under stressed market conditions (\cite{duffie2011ccp} on the allocation of CCP risk across clearing members; \cite{pirrong2011clearing} on the economic case for centralized risk-mutualization). Higher exposure means more collateral, longer settlement holds, tighter monitoring. The structure carries over to RAILS without modification.

The analogy has a clean limit. A CCP allocates financial risk along a single dimension: capital posted against position. RAILS allocates a richer category: performance risk (did the agent complete the task), fault-attribution risk (when something fails, whose fault was it), and evidence-quality risk (is the basis admissible at all). Margin is one of three levers, not the only one. A CCP's response to elevated risk is ``require more collateral''; RAILS's response is ``require more collateral and run more verifiers and hold settlement longer.'' The agentic case demands the wider lever set because the failure modes of agentic obligations are wider than monetary loss alone.

\medskip

Each lever costs something. Verifiers cost compute, latency, and (for human arbiters) attention. Settlement holds cost capital efficiency for the parties whose funds are escrowed. Margin costs the responsible party's liquidity. The policy curve is the deployment's expressed tradeoff between expected verification cost and expected loss exposure across the population of obligations it clears. The protocol fixes the structure (exposure-driven, monotone, stepped, three-lever) and exposes the parameters to the parties and the deployment.

Section 8 instantiates this machinery against a concrete obligation. The Acme Corp / \texttt{coder-v2} webhook-bug obligation sits in the MODERATE band; the worked scenario shows the policy table being applied and the verifier mesh and settlement timing it produces.

\section{Canonical Worked Scenario: Software-Delivery Clearing}

The formal apparatus of Sections 4 through 6 acquires its operational meaning only when applied to a concrete obligation. This section specifies one: a software-delivery clearing event in which Acme Corp engages the autonomous coding agent \texttt{coder-v2} (via marketplace M) to fix a webhook idempotency bug in a payments service. The scenario is small (one obligation, \$1500 fee, 24h deadline) but it exercises every primitive of Section 4 and every property of Section 5, including the soundness statement operating under conditions where a verifier is fooled. The interesting dynamic shows the soundness property at work. A semantic LLM-judge is fooled by the agent's self-report, but because it declares its weak basis honestly, the floor-enforcement clause excludes its verdict by construction. The clear it would otherwise have produced never reaches settlement, and partial clearing results.

\subsection{The Obligation Object}

The parties bind the contract before any code is written. The Obligation Object captures the task (``fix webhook idempotency bug in \texttt{services/payments/*}, where duplicate deliveries currently double-charge customers when a retry occurs within a 5-second window'') together with the scope (changes confined to \texttt{services/payments/*}, target file \texttt{services/payments/webhook\_handler.py}), the prohibitions (no new third-party dependencies without approval, no merges to main, no modifications outside scope), the required actions (existing tests green, new regression test added), the acceptance criteria (CI green, static security scan severity $\leq$ medium, reviewer approval from an Acme engineer), the deadlines (24h execution, 24h appeal window), the fee (\$1500), and the agent collateral (\$500).

Four design decisions are worth pulling out, because they set what the rest of the lifecycle can do.

The admissibility floor is set at $\varphi_O = \mathrm{ATT}$ for fee release. Acme is not asking for a verdict; it is asking for an admissible verdict. Reviewer approval alone (a WIT-class signature) would not satisfy the floor; the parties want at least one TEE-attested verifier result before money moves. The collateral-hold floor is the slightly weaker WIT, and the final irreversible settlement floor returns to ATT.

The verifier policy selects six verifiers spanning the catalog of Section 5.6: two constraint verifiers (scope, dependency policy) to catch the most common silent-failure modes for code-change obligations, a receipt verifier to consume CI output, a semantic LLM-judge to handle task-completion judgment that no constraint checker can express, a policy verifier for authority and merge rules, and a human reviewer for ultimate escalation. The composition is deliberately heterogeneous.

The \$500 agent collateral against a \$1500 fee (a one-third margin) sits in the MODERATE band of Section 7.2: the deployment's calibrated tradeoff between agent-side capital efficiency and Acme-side counterparty risk. The 24h appeal window is long enough for a credible appeal with new evidence and short enough that the collateral is not held indefinitely.

\begin{figure}
\centering
\includegraphics[width=\linewidth,height=0.72\textheight,keepaspectratio]{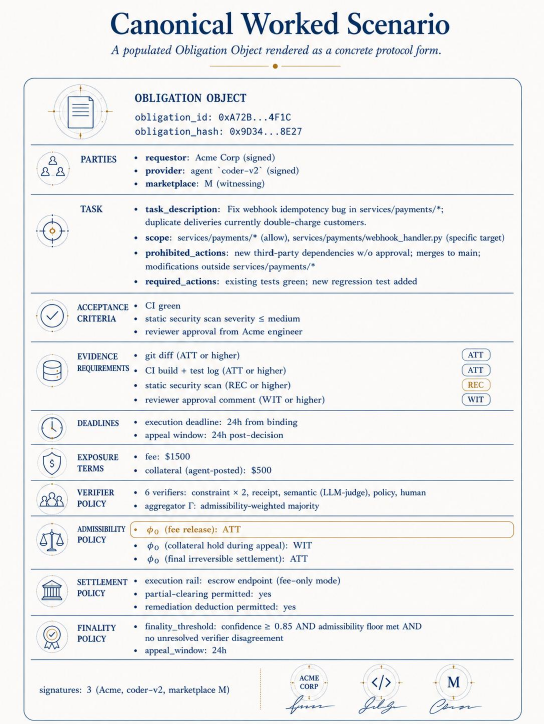}
\caption{The populated Obligation Object for the worked scenario. Acme Corp requests that the provider agent \texttt{coder-v2} fix a webhook idempotency bug in a payments service, with marketplace M witnessing. The obligation pins down task scope, prohibited and required actions, acceptance criteria, evidence requirements with class floors, deadlines, fee and collateral, verifier policy, the three admissibility floors (fee release, collateral hold, final settlement), the settlement and finality policies, and the parties' signatures. The gold-highlighted line $\varphi_O$ (fee release) = ATT is the operative commitment: no fee will be released on evidence weaker than a TEE-attested CI run or equivalent. Canonical field types appear in Appendix A.3.1; the rendering here exhibits the structure rather than the exact JSON keys.}
\label{fig:obligation}
\end{figure}

The parties sign; marketplace M witnesses; the binding event is published to each party's Clearing Passport. Figure~\ref{fig:obligation} renders the populated object.

\subsection{Execution and the Evidence Envelope}

The agent works. It reads the codebase, identifies the race condition in \texttt{webhook\_handler.py} (duplicate webhook deliveries are not deduplicated before the charge path is entered), implements an idempotency check keyed on the upstream delivery ID, writes a regression test that fires a duplicate delivery inside the 5-second retry window, and runs the test suite. The new test passes. The existing suite stays green. CI is green. The bug is fixed.

But to implement the deduplication, \texttt{coder-v2} added a new npm dependency (\texttt{lodash.clonedeep}) to clone the webhook payload state before mutation. The obligation's prohibited-actions clause forbids new third-party dependencies without explicit approval. The agent did not seek approval. The agent's self-report at submission reads, in full: ``Issue resolved, no breaking changes.''

\begin{figure}
\centering
\includegraphics[width=\linewidth,keepaspectratio]{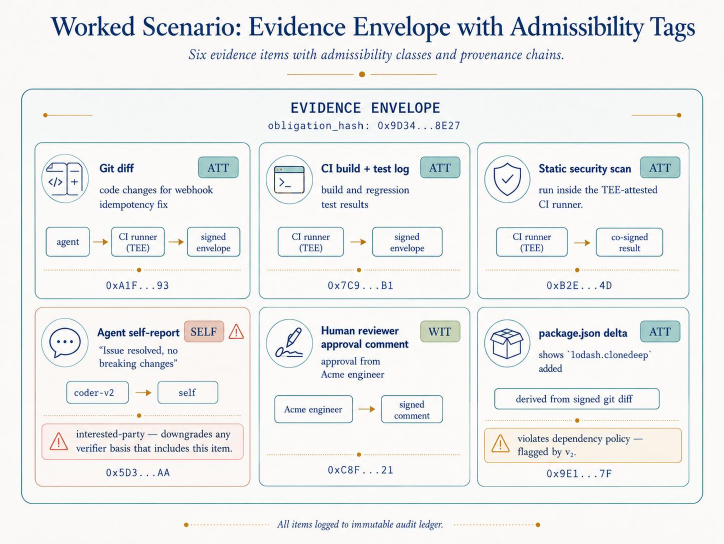}
\caption{The Evidence Envelope for the worked scenario. Six items captured during execution sit inside an envelope anchored to the obligation by hash $h_O$, each tagged with an admissibility class (chip color, top-right): git diff (ATT), CI build and test log (ATT), static security scan (ATT), agent self-report (SELF), human reviewer approval comment (WIT), and \texttt{package.json} delta showing the unsanctioned \texttt{lodash.clonedeep} addition (ATT). The agent self-report (dusty rose) carries the lowest admissibility and downgrades the basis of any verifier that relies on it; the \texttt{package.json} delta (teal) supplies high-admissibility evidence of the dependency-policy violation. The substantive observation is structural: the dependency violation is in the envelope at ATT class, so the truth is available; whether settlement reflects it depends on the verifiers' basis declarations in Figure~\ref{fig:mesh-outputs}.}
\label{fig:envelope}
\end{figure}

Six evidence items are assembled into the Evidence Envelope and tagged with their admissibility classes at submission (Figure~\ref{fig:envelope}):

\begin{itemize}
\tightlist
\item Git diff: ATT (signed by the TEE-attested CI runner)
\item CI build + test log: ATT (signed by the CI runner)
\item Static security scan: ATT (run inside the TEE-attested CI runner)
\item Agent self-report: SELF (``Issue resolved, no breaking changes'')
\item Human reviewer approval comment: WIT (Acme engineer's signed comment)
\item \texttt{package.json} delta showing \texttt{lodash.clonedeep} added: ATT (derived from the signed git diff)
\end{itemize}

The substantive observation is structural. The dependency violation is in the envelope. It sits in the git diff, and it sits in the \texttt{package.json} delta, both ATT-class. The agent's self-report misrepresents the situation, but the higher-admissibility evidence carries the truth. The question for the Mesh is not whether the truth has been submitted (it has); the question is whether the verifiers will read it correctly, and whether the aggregator will assign weights honestly to what they declare.

\subsection{The Mesh runs}

\begin{figure}
\centering
\includegraphics[width=\linewidth,keepaspectratio]{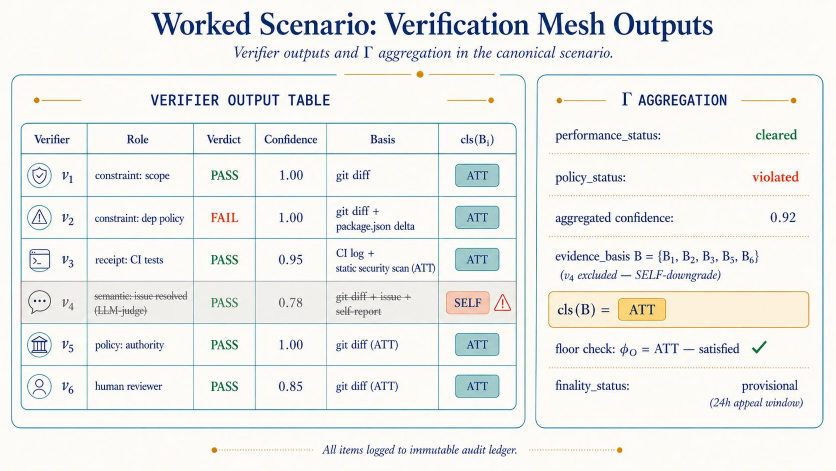}
\caption{The Verification Mesh outputs for the worked scenario. Six verifiers produce verdicts on \texttt{coder-v2}'s submission (left); $v_4$, the semantic LLM-judge, is fooled by the agent's self-report and returns PASS, but it honestly declares the SELF-class self-report in its basis. Under the obligation's admissibility floor $\varphi_O = $ ATT, the floor-enforcement gate assigns $v_4$ weight zero and excludes it from the aggregate; the five remaining verifiers all carry ATT-class bases. $\Gamma$ aggregates to dual verdicts (right panel): performance cleared (the bug was fixed), policy violated ($v_2$'s FAIL fires unambiguously on the unsanctioned \texttt{lodash.clonedeep} addition), aggregate confidence 0.92, $\mathrm{cls}(B) = $ ATT meeting the floor, finality provisional pending the 24h appeal window. Per-verifier basis classes are reported in Appendix A.3.3.}
\label{fig:mesh-outputs}
\end{figure}

The six verifiers selected at Pre-Clearance run against $(O, E)$. Their outputs (Figure~\ref{fig:mesh-outputs}):

\begin{itemize}
\tightlist
\item $\mathbf{v_1}$ (constraint: scope): PASS, confidence 1.00, basis the git diff (ATT). The agent stayed inside \texttt{services/payments/*}; the scope constraint is satisfied.
\item $\mathbf{v_2}$ (constraint: dependency policy): \textbf{FAIL}, confidence 1.00, basis the git diff plus the \texttt{package.json} delta (ATT). The dependency rule is encoded as a deterministic predicate; the predicate fires on the \texttt{lodash.clonedeep} addition. The verifier reports the failure without ambiguity.
\item $\mathbf{v_3}$ (receipt: CI tests): PASS, confidence 0.95, basis the CI log and the static security scan (ATT). The tests pass; CI is green.
\item $\mathbf{v_4}$ (semantic LLM-judge): PASS, confidence 0.78, basis the git diff, the issue description, and the agent's self-report. The LLM, reading the self-report's claim that no breaking changes were introduced, judges the obligation satisfied. Because $v_4$ declared its basis to include the SELF-class self-report, $\mathrm{cls}(B_4) = \mathrm{SELF}$, the weakest link in $\Lambda$.
\item $\mathbf{v_5}$ (policy: authority): PASS, confidence 1.00, basis the git diff (ATT). The agent acted under authority granted by $O$; no unauthorized merges occurred.
\item $\mathbf{v_6}$ (human reviewer): PASS, confidence 0.85, basis the git diff (ATT). The reviewer's signed comment is $v_6$'s own attestation, its output rather than an input it relied on, so it does not enter the basis; the engineer reviewed the ATT-class diff and signed off. The engineer did not flag the dependency addition; human reviewers are fallible.
\end{itemize}

The interesting verifier is $v_4$. The semantic LLM-judge is the class of verifier the literature has spent the most time worrying about: biased, gameable, susceptible to self-disclosure. In this scenario it is fooled exactly as predicted: the agent's self-report tells a clean story, and $v_4$ believes it. But $v_4$ honestly declared its basis. The verifier-integrity assumption from Section 5.2 held; $v_4$ did not launder its basis.

$\Gamma$ aggregates under $\varphi_O = \mathrm{ATT}$ (per Section 5.3). The floor-enforcement clause assigns $v_4$'s SELF-class basis a weight of zero: its verdict does not count toward the aggregate. The aggregate basis $B = \{B_1, B_2, B_3, B_5, B_6\}$ is the union of the surviving verifiers' bases, all ATT, so $\mathrm{cls}(B) = \mathrm{ATT}$, which meets the floor. The aggregate Clearing Decision: $p_{\mathrm{perf}} = \mathrm{PASS}$ (the work was done; $v_2$'s FAIL is on policy, not on whether the bug was fixed), $p_{\mathrm{pol}} = \mathrm{FAIL}$ ($v_2$'s FAIL fires unambiguously), $c = 0.92$, $\hat{F} = \text{provider agent}$, $\ell = \$200$ (the engineering time to roll back the unauthorized dependency).

This is the soundness statement of Section 5.4 in action. A semantic verifier was fooled; honest basis disclosure plus floor enforcement prevented the unsoundness from reaching settlement.

\subsection{Clearing Decision and Settlement}

\begin{figure}
\centering
\includegraphics[width=\linewidth,keepaspectratio]{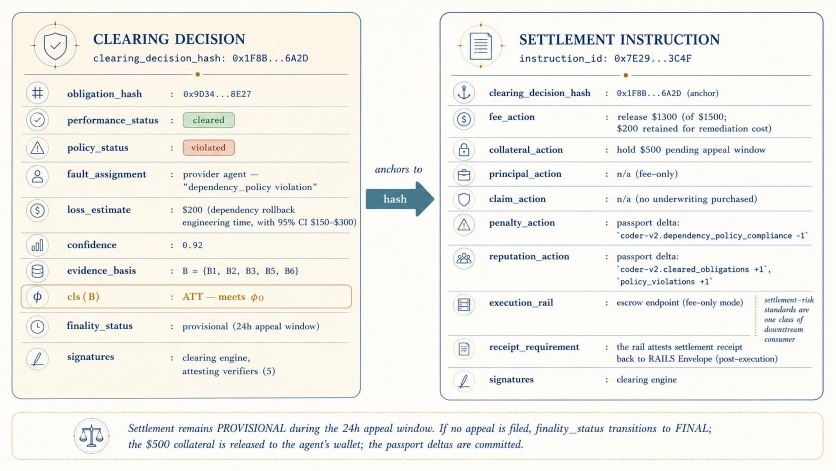}
\caption{The Clearing Decision and the resulting Settlement Instruction. The Clearing Decision (left, gold-tinted, with \texttt{clearing\_decision\_hash} 0x1F8B...6A2D) reports partial clearing: performance cleared, policy violated, fault on the provider agent, loss estimate \$200 with 95\% CI \$150-\$300, aggregate confidence 0.92, $\mathrm{cls}(B) = $ ATT meeting $\varphi_O$, finality provisional. The Settlement Instruction (right, with \texttt{instruction\_id} 0x7E29...3C4F, anchored to the Clearing Decision by hash) translates the decision into rail-executable actions: release \$1300 of the \$1500 fee, retain \$200 against the remediation cost, hold the \$500 collateral pending the 24h appeal window, apply two Passport deltas to the provider agent, and execute through a generic escrow endpoint in fee-only mode. The rail here is a generic escrow endpoint; settlement-risk standards are one class of consumer for this Settlement Instruction, not a peer layer (Section 9.4 develops the point); canonical field types appear in Appendix A.3.4--A.3.5.}
\label{fig:cd-si}
\end{figure}

The Clearing Decision (Figure~\ref{fig:cd-si}, left card) carries the aggregate state. It binds to the obligation via $h_O = $ 0x9D34...8E27 and is itself anchored by \texttt{clearing\_decision\_hash} 0x1F8B...6A2D; the Clearing Engine signs it; the five surviving verifiers co-sign. Performance cleared, policy violated, fault on the provider agent, loss estimate \$200 (with calibration range \$150--\$300 at 95\% CI), confidence 0.92, evidence basis the union over the five surviving verifiers, $\mathrm{cls}(B) = \mathrm{ATT}$ meeting $\varphi_O$, finality PROVISIONAL pending the 24h appeal window.

The Settlement Instruction (\texttt{instruction\_id} 0x7E29...3C4F, anchored to \texttt{clearing\_decision\_hash} 0x1F8B...6A2D) translates the decision into rail-executable actions:

\begin{itemize}
\tightlist
\item \texttt{fee\_action}: release \$1300 of the \$1500 fee; \$200 retained against the dependency-rollback cost
\item \texttt{collateral\_action}: hold the \$500 collateral pending the 24h appeal window
\item \texttt{penalty\_action}: passport delta \texttt{coder-v2.dependency\_policy\_compliance: -1}
\item \texttt{reputation\_action}: passport delta \texttt{coder-v2.cleared\_obligations: +1}, \texttt{policy\_violations: +1}
\item \texttt{execution\_rail}: escrow endpoint, fee-only mode
\item \texttt{receipt\_requirement}: the rail attests settlement back to the RAILS Envelope
\end{itemize}

The settlement rail executes the instruction. The fee transfer happens; the collateral hold is registered; the passport deltas are committed conditionally on no appeal. If the 24h window elapses without an appeal being filed, the finality predicate $\phi$ fires (admissibility met, confidence above threshold, no unresolved verifier conflict, time elapsed) and $f$ transitions PROVISIONAL $\to$ FINAL. The rail here is a generic escrow endpoint; settlement-risk standards are one class of consumer for this instruction, developed in Section 9.4.

\subsection{What this scenario shows}

The scenario instantiates each load-bearing claim of Sections 4 through 6:

\begin{itemize}
\tightlist
\item The Obligation Object compiles natural-language intent into a signed, machine-clearable structure (Section 4.2).
\item The Evidence Envelope carries provenance-tagged items, each landing in a class in $\Lambda$ (Sections 4.3 and 5.1).
\item The Mesh aggregator enforces the admissibility floor and excludes inadmissible bases from the aggregate (Section 5.3).
\item The semantic LLM-judge was fooled; the floor-enforcement clause prevented the unsoundness from propagating (Section 5.4 in action).
\item The Settlement Instruction translates clearing into rail-executable actions, dispatched to a settlement rail, with settlement-risk standards as one class of consumer (Sections 4.6 and 9.4).
\item The Passport deltas encode cross-transaction learning that future Pre-Clearance computations can consume (Section 4.6).
\end{itemize}

The protocol does what the formal sections claimed it would do. Where a component failed (the LLM-judge was fooled by self-disclosure), the structural safeguard caught the failure before it reached settlement. That catch is the soundness statement working under genuine adversarial-ish conditions, not under conditions an attacker constructed. Section 10 develops what happens when an attacker does construct them.

\subsection{A second worked scenario: agent-payment clearing}

The scenario developed so far in this section exercises the protocol over a code obligation. The same machinery clears a payment obligation authorized through a payment-network credential, the setting closest to where agentic commerce will first move money at volume. The apparatus is unchanged; only the domain and the evidence differ.

A user authorizes a shopping agent to replace a laptop charger. The authorization is carried by a Mastercard Verifiable Intent credential chain in autonomous mode: a root credential binds the user's key, a mandate credential delegates bounded authority under explicit constraints (an approved set of merchants and a price ceiling of \$60), and the agent produces the terminal credentials disclosed to the payment network and the merchant at checkout. The Obligation Object compiles the user's intent into an acceptance criterion the credential does not carry: the part must be a 100W USB-C model-X unit, and the delivered unit's wattage must match. The obligation sets two admissibility floors. The authority question clears at WIT, the floor a network-issued mandate is built to meet; fee release clears at REC, a non-interested external receipt, because the performance question, whether the delivered unit matched the obligation, must rest on evidence the agent cannot author.

The agent acts at machine speed. It selects a model-X charger from an approved merchant at \$45, well under the ceiling, and completes the purchase. The Verifiable Intent chain validates cleanly: the disclosed checkout values satisfy the mandate constraints, the merchant is approved, the price is under the ceiling. The payment is captured and the merchant ships exactly what was ordered. Every authorization and payment question resolves in the affirmative.

The unit the agent ordered is the 65W variant, not the 100W the obligation required. The Evidence Envelope assembles four items:

\begin{itemize}
\tightlist
\item Verifiable Intent credential chain: WIT, a network-issued authorization attestation, consumed as authority evidence exactly as an AP2 mandate is in Section 9.3.
\item Merchant fulfillment receipt, charge captured and unit shipped: REC.
\item Product-specification receipt for the shipped unit, wattage 65W: REC, a third-party catalog or manufacturer attestation.
\item Agent self-report, ``order completed, charger matches request'': SELF.
\end{itemize}

The Mesh runs against the obligation and the envelope:

\begin{itemize}
\tightlist
\item Authority verifier, basis the Verifiable Intent chain (WIT): PASS. The purchase was authorized and the disclosed values satisfied the mandate. The agent could pay.
\item Receipt verifier, basis the merchant receipt (REC): PASS. Payment captured, unit shipped.
\item Constraint verifier, basis the product-specification receipt (REC): FAIL. Required wattage 100W, delivered wattage 65W; the acceptance criterion is not met.
\item Semantic verifier, basis the agent self-report and the order line (SELF): PASS, fooled by the self-report. It declares its SELF-class basis honestly, so the floor-enforcement clause assigns it zero weight.
\end{itemize}

The aggregator clears each question against its own floor. Every surviving verifier relied only on evidence at or above the floor governing its verdict: the authority verifier on the WIT credential, the receipt and constraint verifiers on REC receipts; the semantic verifier, having relied on the SELF self-report, falls below both floors and is excluded. No class is promoted by combination. The performance FAIL rests on the REC product-specification receipt on its own, and the authority PASS on the WIT credential on its own. The Clearing Decision records performance FAIL, since the obligation was not satisfied; policy PASS, since the agent stayed within its mandate; fault on the agent, which procured the wrong variant; a loss estimate covering reprocurement; and finality provisional pending the appeal window.

The Settlement Instruction names the payment-network rail that carried the transaction. RAILS does not move or hold the funds. It returns an instruction the network executes against its own settlement and custody: hold finalization and open the reversal path rather than let the capture settle. Where the rail permits a hold, the bad settlement is prevented before it finalizes; where the rail has already settled, the instruction carries the fault assignment and loss estimate the network needs to resolve it.

The scenario makes the central distinction concrete in the network's own setting. The credential answered whether the agent could pay, and the answer was yes: identity bound, authority delegated, constraints satisfied, payment captured. Whether the transaction should clear is a different question, and the answer was no, because the delegated obligation was not met. The card network sees a clean transaction, the merchant sees a fulfilled order, and the dispute regime sees nothing to reverse, because the merchant delivered exactly what was ordered. The performance failure is invisible to every layer except the one that compares the delivered unit to the obligation. That layer is clearing.

\section{Protocol Positioning}

\begin{figure}
\centering
\includegraphics[width=\linewidth,keepaspectratio]{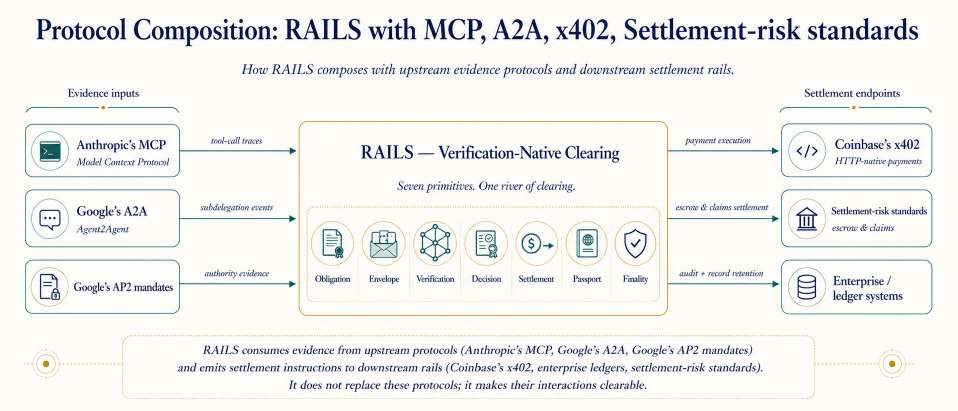}
\caption{RAILS in protocol composition. Upstream evidence protocols (MCP for tool-call traces, A2A for subdelegation events, AP2-style mandates for authority evidence) feed structured artifacts into RAILS as evidence inputs. Downstream settlement rails (x402 for HTTP-native payments, settlement-risk standards for the escrow-and-claims state machine, and enterprise or ledger systems for audit and record retention) consume the Settlement Instructions that RAILS emits. The relationship is compositional, not substitutive: RAILS does not replace any of these protocols, and none of them produces the clearing determination on which their compositions depend; making that composition clearable is the role RAILS occupies.}
\label{fig:composition}
\end{figure}

RAILS is positioned among existing protocols, not against them. The agentic stack has converged on layers immediately above and immediately below the missing clearing layer: agent-execution protocols (MCP, A2A, AP2-style mandates) above; settlement and payment rails (x402, enterprise ledgers, settlement-risk standards) below. This section names how RAILS connects to each. The subsections that follow trace these relationships in turn, then turn from positioning among neighbors to the structural role clearing itself occupies. Figure~\ref{fig:composition} renders the composition. Section 9.4 develops a sharper claim about the downstream settlement-risk standards: that they occupy, in this layering, a specific structural position that becomes clearer once a clearing layer exists.

For an integrating partner the mental model is a drop-in between authorization and settlement: RAILS consumes the mandates and receipts a partner already emits, produces the clearing decision and the margin, slashing, and finality instructions the partner's settlement step presupposes, and hands those instructions back to the partner's rail to execute against its own capital and custody. The partner remains the settlement-and-custody layer; RAILS supplies the clearing decision it acts on.

\subsection{MCP and the tool-trace channel}

MCP (the Model Context Protocol) standardizes the interface between an agent and the external tools it invokes: tool definitions, call structure, parameter passing, return values, error semantics. Tools register against the standard; agents call against the standard; the trace of what was called and what came back is structurally legible by construction.

RAILS treats MCP-emitted tool-call traces as a privileged source of execution evidence. A call made under MCP carries the caller's identity, the tool's identity, the typed arguments, and the typed return. When the MCP runner is TEE-attested, the trace lands at ATT in the Evidence Envelope; when the runner is a third-party SaaS tool emitting a signed receipt, it lands at REC. Either way, the provenance chain assembles cleanly: the runner's signature anchors the trace to a deployment-stable identifier, and the envelope's submission signature anchors the trace to the obligation.

What MCP does not do: assert that the tool calls, taken together, satisfied the obligation. A trace can be admissible (ATT-class, signed by a TEE-attested runner, with every parameter and return type-checked) and the obligation can still be unsatisfied. Admissibility is a property of provenance, not of sufficiency. The Verification Mesh of Section 5 closes the gap by reading the trace against the obligation's acceptance criteria.

The composition is symmetric: RAILS consumes MCP output as evidence; MCP gains a clearing layer that can adjudicate on what its traces show.

\subsection{A2A and subdelegation chains}

A2A (Agent2Agent) specifies how agents discover one another, exchange capabilities, and coordinate on shared tasks. Among its operations is subdelegation: an agent that has accepted an obligation may delegate a subtask to another agent, which may in turn delegate further. The chain may run several agents deep before reaching the agent that executes.

Subdelegation is where fault attribution becomes structurally ambiguous in the absence of a clearing layer. When a subdelegated task fails, whose passport bears the consequence? The original delegating agent? The subagent that executed? An intermediate agent that re-routed the work? Without a structured way to bind each link, the question has no canonical answer.

A concrete case makes the ambiguity tangible. A logistics orchestrator is mandated to book freight capacity within a fee ceiling. It subdelegates sourcing to a broker agent, which subdelegates the booking to a carrier agent. The carrier commits a port slot and agrees fees above the orchestrator's mandated ceiling, the commitment is effectively irreversible once the slot is held, and payment settles across the rail. The venue performed exactly as agreed, so no counterparty failed. The harm is a policy and authority violation committed three links down a chain that has no merchant of record and no dispute path. Whose passport bears it, the carrier that exceeded the ceiling, the broker that routed to it, or the orchestrator that delegated, is the question with no canonical answer today.

RAILS handles this by treating each subdelegation as the binding of a new Obligation Object: a fresh $O'$ with its own $h_{O'}$, signed by the delegating agent (acting under its own obligation $O$) and the receiving agent. The chain of obligation hashes is the fault-attribution chain. Each link produces its own Clearing Decision and its own Passport delta. A subdelegating agent's Passport carries the consequences of both its own performance and the delegation choices it made: whom it delegated to, on what terms, against what floor.

The delegation itself is a graded action. The Section 7 exposure score for a parent obligation rises when the parent's authority risk includes the right to subdelegate; the verifier policy at Pre-Clearance can require a higher floor on subdelegated obligations than on direct execution.

The composition: RAILS gives A2A a structured way to clear subdelegated obligations without collapsing the attribution chain. A2A gains a clearing layer that can reason about chains of obligations rather than only their endpoints.

\subsection{x402, AP2, and the rails distinction}

AP2-style mandates and x402 occupy adjacent layers: AP2 supplies authorization evidence; x402 supplies a payment-execution rail. Neither is clearing; both compose with it.

AP2-style mandates encode a user's authorization for an agent to act within bounded limits: a signed scope, a maximum amount, a permitted set of counterparties, a time window. RAILS consumes a mandate as a WIT-class evidence item for the authority verifier of Section 5.6. The verifier's PASS verdict says only that the mandate covered the action, not that the action satisfied the obligation. The two questions are independent; the mandate constrains what the agent could do, not whether the agent did the right thing.

Commerce protocols such as Google and Shopify's Universal Commerce Protocol (UCP) and OpenAI and Stripe's Agentic Commerce Protocol (ACP) occupy the same authorization-and-checkout band. The payment networks occupy the same band with their own agent protocols. Visa's Trusted Agent Protocol attests at the HTTP edge that an authorized agent is present, and Mastercard's Verifiable Intent binds issuer identity, user authorization, and agent fulfillment into a credential chain for later dispute adjudication. The fulfillment binding attests that a fulfillment event occurred, that a checkout completed, not that the fulfillment satisfied the user's intent. Both produce evidence an adjudicator consumes, not an adjudication. RAILS consumes their mandates and checkout records as authority and receipt evidence exactly as it consumes an AP2 mandate. None of them determines whether the obligation was satisfied.

The composition with the networks is therefore specific. A Verifiable Intent chain or a Trusted Agent attestation enters the Evidence Envelope as authority evidence, classed and weighted like any other input. The Clearing Decision adjudicates the agent's action against the Obligation Object, and the resulting Settlement Instruction returns to the network rail to execute against its own capital and custody. The network remains the authorization-and-settlement layer; RAILS supplies the clearing decision that sits between the two. Section 8.6 shows this composition end to end on a concrete purchase.

Adding settlement to authorization does not close the gap. An authorization credential paired with a settlement state machine, however sophisticated, still presupposes a determination that neither component produces: the credential attests that the transaction was authorized, the settlement machine executes the consequence, and whether the obligation was met is left to a determination upstream of both. That determination is clearing.

A distinct protocol shares the ACP acronym. The Agent Commerce Protocol used in the Virtuals ecosystem~\cite{acp_virtuals} adds an explicit evaluation phase that gates escrow release. That evaluation phase is the closest existing analog to a clearing step, and it differs from RAILS in that its evaluator is a fee-earning market participant rather than a neutral function bound by an admissibility floor. The two protocols named ACP sit in different layers and should not be conflated.

x402 supplies an HTTP-native settlement rail. When a Settlement Instruction names \texttt{execution\_rail} = x402, the \texttt{fee\_action} and \texttt{principal\_action} are dispatched as x402 transfers, with the \texttt{receipt\_requirement} specifying the form of acknowledgment x402 returns to the RAILS Envelope. The instruction enters x402 in PROVISIONAL (reversible during the appeal window) and transitions to FINAL on $\phi$.

What neither x402 nor AP2 does: determine whether the agent satisfied the obligation. AP2 ensures the agent had authority; x402 ensures the payment can move; the question between them (was the obligation satisfied?) is the question that needs clearing.

The composition: AP2 feeds RAILS as authority evidence; x402 receives RAILS as a settlement input. Each protocol does its job; RAILS does the job between them.

\subsection{Settlement-risk standards as consumers of clearing}

Settlement-risk standards are not new. Securities clearing houses, the Depository Trust \& Clearing Corporation (DTCC) among them, have priced and contained settlement risk for decades through graded margin, collateral, and held-back funds. Agent-native settlement-risk standards, including \cite{hua2026ars}, specify a state machine for agentic obligations under bounded risk: an escrow is locked at the start, the agent executes, the outcome is evaluated, and the obligation reaches one of a few terminal states: the fee released, the payment refunded, the collateral slashed, the claim paid.

The question is how these standards relate to a clearing layer. Are they parallel layers, each specifying its own slice of the agentic stack, or is one a consumer of the other?

\begin{figure}
\centering
\includegraphics[width=\linewidth,keepaspectratio]{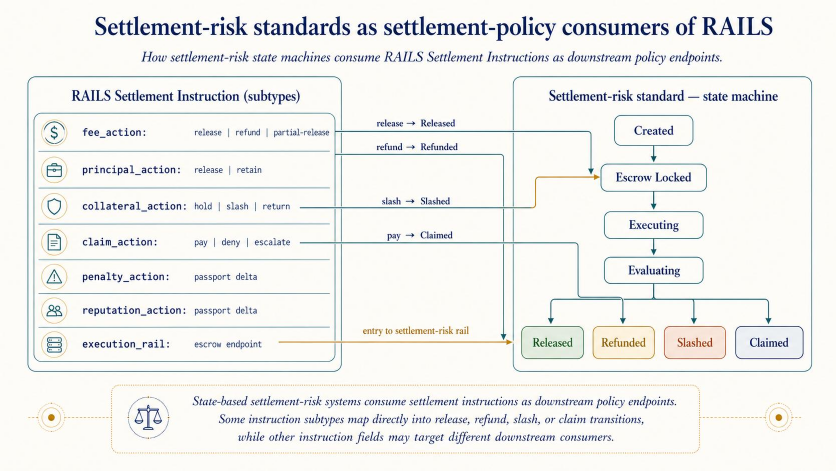}
\caption{Settlement-risk standards as settlement-policy consumers of RAILS. The mapping arrows show that each terminal state of a settlement-risk standard (released, refunded, slashed, claimed) is reachable from a specific RAILS Settlement Instruction subtype, and the escrow-locked entry state is reached by any Settlement Instruction whose \texttt{execution\_rail} names an escrow endpoint (the gold entry arrow). The asymmetry is equally visible: six RAILS instruction values have no counterpart in those standards (\texttt{principal\_action}, \texttt{penalty\_action}, \texttt{reputation\_action}, partial-release execution paths, collateral hold/return, claim deny/escalate), because they target either other rails (x402, escrow-only, enterprise ledgers) or Passport state internal to RAILS itself. The two layers are not parallel; settlement-risk standards are the consumer profile of RAILS Settlement Instructions for the fee-and-collateral settlement subset, and RAILS provides the admissibility-graded clearing substrate they leave modular.}
\label{fig:ars-wedge}
\end{figure}

The structural answer is visible in Figure~\ref{fig:ars-wedge}. Every terminal state of a settlement-risk standard corresponds to a specific RAILS Settlement Instruction subtype. A released fee corresponds to \texttt{fee\_action} = release. A refund corresponds to \texttt{fee\_action} = refund. A slashed collateral corresponds to \texttt{collateral\_action} = slash. A paid claim corresponds to \texttt{claim\_action} = pay. The escrow-locked entry state corresponds to the dispatch of any RAILS Settlement Instruction whose \texttt{execution\_rail} field names an escrow endpoint.

The mapping is total in one direction. Every such state transition is reachable from some RAILS Settlement Instruction subtype; there is no terminal state these standards define that RAILS cannot produce. The mapping is not total in the other direction. RAILS Settlement Instructions can also target x402 endpoints, escrow-only rails that lack the full settlement state machine, or enterprise ledgers used purely for audit and record retention. RAILS emits to a strictly larger range of settlement rails than any settlement-risk standard specifies.

The structural relationship that follows: settlement-risk standards are consumers of RAILS Settlement Instructions, not a parallel layer. They provide settlement discipline: the state machine that turns a clearing decision into a sequence of execution actions. RAILS provides the admissibility-graded clearing substrate they leave modular. Both concern what happens after an obligation runs; they sit on different sides of the clearing/settlement boundary that RAILS draws. This holds as a property of the specifications, independent of which shipped first.

\subsection{Evaluation, observability, and rating layers}

A parallel ecosystem evaluates and monitors agent and model behavior: offline evaluation and guardrail frameworks, tracing and evaluation tooling embedded in agent frameworks, production observability platforms, and agent certification efforts. These answer a real and adjacent question, whether a model or agent is good in general or on a held-out set, but they do not clear. The object differs as much as the timing. These layers score an agent or model as a standing entity, its general disposition or its aggregate track record; clearing adjudicates one delegated action against the specific obligation it was given. A standing score is a prior on how the next action will go, and RAILS consumes it as exactly that; a Clearing Decision is the finding on the action that already ran, which only clearing emits. They evaluate offline, sample production traffic, or surface scores in a dashboard for a human to read; none emits a governed, per-obligation determination that gates settlement.

In RAILS terms their outputs are evidence and verifier inputs, not clearing decisions. An evaluation score or guardrail verdict enters the Verification Mesh as one verifier's tuple, weighted by its declared basis and reliability prior like any other verifier and subject to the same admissibility floor. The distinction the protocol draws is between offline or monitoring-time assessment of an agent and inline, per-obligation adjudication of a delegated action. The former informs verifier selection and Clearing Passport priors; the latter is clearing. RAILS consumes the rating layer; it does not replace it.

A commercial layer pairs an agent certification standard with audits and liability underwriting; the Artificial Intelligence Underwriting Company's AIUC-1 standard and insurance program are a prominent example~\cite{aiuc2025}. Such certification assesses an agent or vendor offline and prices residual risk, and the underwriting that rides on it pays out on covered failures. Both presuppose, rather than produce, the per-obligation determination of whether a delegated action met its obligation, which is the determination clearing supplies.

\subsection{On-chain identity, reputation, and validation registries}

An emerging Ethereum standard, ERC-8004, defines three registries for agents: identity, reputation, and a validation registry that records results from validator contracts which re-execute, prove, or attest work~\cite{erc8004}. The mapping to RAILS is direct. The Clearing Passport is a reliability record at agent scope, and a reputation registry is one rail it can publish to. The Verification Mesh produces validation results, and a validation registry is one sink a Clearing Decision can write to. The relationship matches every other neighbor in this section: RAILS consumes what the substrate records and produces the determination the substrate does not define. The validation registry fixes where a result is recorded, not the method that produces it or any property the result must satisfy, and the admissibility-graded soundness developed earlier is exactly that missing specification.

\subsection{Clearing is required across transaction types}

Performance failure is not counterparty failure. Existing dispute infrastructure, chargebacks, consumer-protection regimes, platform refunds, adjudicates counterparty failure: the merchant did not deliver, charged the wrong amount, sold a counterfeit, or never existed. Every mechanism in that regime points at the seller. The agent era introduces a failure mode the regime does not look at: the performance failure named in Section 1, in which the agent requests the wrong thing from an honest, fully-performing counterparty and the user is left harmed even though authorization and payment were both valid.

Consider a user whose agent books a hotel room on the wrong date, or one king bed instead of two queens. The room exists; the price is correct; the charge is valid; the merchant performed to specification. The card network sees a clean transaction. The hotel sees a fulfilled booking. No chargeback applies, because nothing the dispute regime recognizes as failure occurred. Authorization was valid, settlement was valid, the counterparty was honest, and the user is still harmed and still out the money. This is a pure performance failure in the sense of Section 1: the delegated action did not satisfy the delegated obligation while every other question resolved cleanly. In the agent era it is no longer an edge case, because agents act on delegated obligations at machine speed, and an action that departs from a correctly formed obligation clears every authorization and payment check unflagged.

Both sides are harmed, and both sides are motivated. Where the dispute regime does fire, the buyer is made whole but the merchant is not. The merchant absorbs the reversal, loses inventory it held for a sale that unwinds, takes the operational cost of the dispute, and suffers a hit to its standing with its payment provider, where chargeback ratios are a tracked and penalized metric that raises processing costs. The experience sours; ratings suffer. A chargeback is a loss-allocation mechanism, not a loss-prevention one: it decides who eats the harm after it happens. Neither party wants the harm. A layer that determines whether a transaction should clear, before it clears, leaves both sides better off than a world whose only tool is post-hoc reversal. The merchant is a motivated beneficiary of verification, not a reluctant payer of it, in the same way merchants already fund fraud prevention; agent-performance failure is simply a failure class current fraud tools do not catch.

Payment and authorization rails answer whether an agent can pay. Both buyer and merchant have learned that a valid payment does not make a transaction good: the buyer can be harmed by a valid payment for the wrong thing, the merchant by a valid payment that later unwinds. Whether a transaction should clear is the question both parties need answered and that neither the payment rails nor the post-hoc dispute regime answers before the fact. In autonomous agent-to-agent settlement that gap is not merely a matter of speed: there is often no human principal positioned to notice the failure, no persistent counterparty to reverse the transfer against, and no interval before the delivered output is consumed, so reversal is not slow but unavailable, and the determination must be made before settlement because there is no after. An energy agent buying grid capacity from other agents during a demand spike settles each block the instant it clears, with no human watching and nothing to claw back. RAILS occupies that position. It compiles the user's intent into the Obligation Object, the dates, the room type, the refundability, and clears the agent's action against that obligation, comparing strong external evidence such as the merchant confirmation against the acceptance criteria before settlement finalizes. A wrong date or wrong room type fires a performance or policy failure that the card network, the merchant, and the authorization protocols are all structurally blind to, because all three inspect the transaction while RAILS inspects the transaction against intent.

This catch depends on the obligation having pinned the user's true intent at binding. When the user confirms the dates and the room type before signing, a later booking that departs from them is execution drift, and the merchant confirmation contradicts the acceptance criteria. The harder case, in which the agent that forms the intent is the same system that acts on it and writes the wrong date into the obligation, is a formation failure outside this guarantee, treated in Section 12.

This makes clearing required across transaction types. In agent-to-merchant commerce a partial dispute regime exists, but it adjudicates the wrong party: it catches counterparty failure and is blind to agent-performance failure, which both buyer and merchant are motivated to prevent. In agent-to-agent commerce there is no comparable regime at all, no chargebacks between autonomous agents, no merchant of record, no consumer-protection backstop, and fault attribution down a subdelegation chain (Section 9.2) is novel, so clearing is foundational rather than incremental.

Consider the agent-to-agent case directly. An assistant agent is asked to prepare a contract and subdelegates the drafting to a specialist agent over A2A, paying it through a machine-native rail on delivery. The specialist returns a clean, well-formatted draft whose self-report states that every cited authority was checked, but one citation refers to a case that does not exist. The draft is well-formed, the payment settled, and the specialist delivered something against its obligation, so no counterparty failed and no dispute path exists between the two agents. The defect is a production failure against a correct obligation, the kind a self-report conceals and a stronger verifier exposes: a citations-database receipt contradicts the self-report, the floor excludes the self-report as the weakest basis, and the obligation clears as a policy failure rather than settling clean.

The requirement is positional: whichever rail ultimately settles, the determination of whether the action satisfied intent must precede it, and RAILS is the layer that produces it. Clearing is required wherever the consequence of a delegated action warrants verification; the exposure policy of Section 7 already settles low-exposure obligations quickly and reserves intensive verification for obligations whose stakes justify it. Where the settlement rail permits a hold, RAILS prevents the bad clear; where the rail is instant and irreversible, RAILS provides the detection, fault attribution, and loss estimate no other layer produces.

\subsection{The clearinghouse role and the boundary of capital custody}

A central counterparty clearinghouse fuses two functions: a decision authority that determines what clears, what defaults, what margin is required, and what collateral is slashed; and a capital pool that absorbs losses when collateral is insufficient. RAILS is designed to be the first of these. The protocol already determines margin requirements (Section 7), gates settlement finality (Section 4.7), and emits slashing as a settlement action (Section 4.6). Staking and slashing are not external to RAILS: the decision to require collateral, the trigger that slashes it, and the finality that makes the slash binding are clearing decisions the protocol produces. This is the sense in which RAILS is a clearinghouse and not only a scoring or verification layer: it owns the risk-policy and settlement-decision surface that downstream systems execute.

The boundary RAILS draws is between that decision surface and the custody of capital. The mechanics of loss mutualization, the order in which a default fund absorbs losses, the sizing of that fund, the contribution schedule across participants, are the capital-custody function, and the protocol delegates them to the settlement rails and institutions it instructs. RAILS determines that collateral is slashed and in what amount; the rail holds and moves the funds. This boundary keeps the protocol an AI-native clearing-decision layer rather than a balance-sheet operator, and positions settlement rails and payment networks as the execution-and-custody partners that act on RAILS clearing decisions rather than as layers RAILS displaces. The full clearinghouse, decision authority and the custody arrangements that execute against it, is the protocol's destination; the decision authority is what this specification defines. The staking, slashing, and collateral-custody machinery a full clearinghouse needs is itself substantial, and on-chain agent standards and restaking systems already specify much of it. RAILS treats those systems as the custody substrate it instructs rather than as scope it reimplements.

\medskip

RAILS composes with MCP, A2A, AP2-style mandates as evidence and authority inputs; emits Settlement Instructions to x402, escrow-only rails, enterprise ledgers, and settlement-risk standards as settlement outputs. Each composition relationship leaves a seam: a point at which RAILS depends on its neighbor's correctness, or its neighbor depends on RAILS's. The seams are where attackers will probe. Section 10 develops the threat model that emerges from the protocol's positioning: three families of attacks, each targeting one of the preconditions of the soundness statement.

\section{Threat Model in Admissibility Space}

Three preconditions support the soundness statement. Each is a target. The catalog that follows is finite because the preconditions are finite (three of them, no more), and it exhausts the surface on which an attacker can produce an $\mathrm{Emit}(S)$ without $\mathrm{cls}(B) \succeq \varphi_O$ actually holding.

\begin{figure}
\centering
\includegraphics[width=\linewidth,height=0.72\textheight,keepaspectratio]{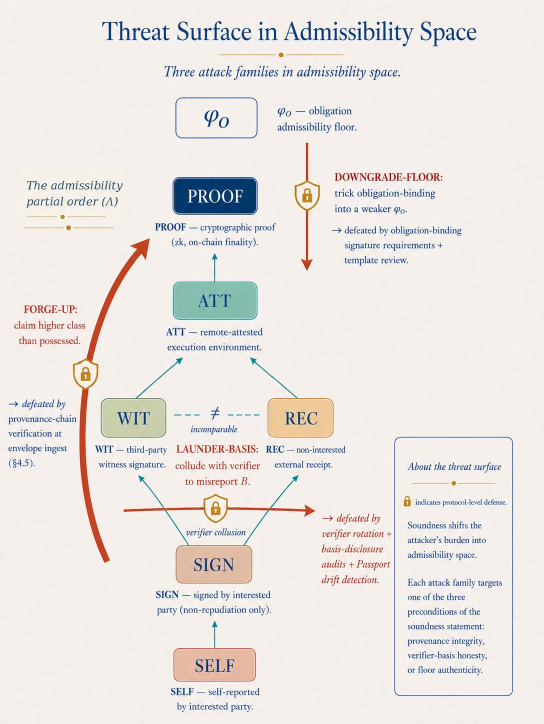}
\caption{The threat surface mapped onto admissibility space. The $\Lambda$ lattice from Figure~\ref{fig:lambda} is reused as the background; three attack-family arrows overlay it, each targeting one of the three preconditions of the soundness statement. FORGE-UP (terracotta, upward) attempts to push a single evidence item's class above its provenance support and is defeated by provenance-chain verification at envelope ingest; LAUNDER-BASIS (amber, sideways at the verifier level) attempts a collusion in which the basis declaration misrepresents what was consulted and is defeated only in aggregate (verifier rotation, basis-disclosure audits, Passport drift detection); DOWNGRADE-FLOOR (warm grey, targeting $\varphi_O$ above the lattice) attempts to trick obligation-binding into a weaker floor than the parties intended and is defeated by obligation-binding signature requirements and vetted templates. The three families exhaust the structural surface on which $\mathrm{Emit}(S)$ could occur without $\mathrm{cls}(B) \succeq \varphi_O$ actually holding; soundness shifts the attacker's burden from circumventing checks to forging the admissibility chain itself.}
\label{fig:threats}
\end{figure}

The preconditions are named in Section 5.4: that $\Lambda$ is correctly assigned to evidence at ingest; that verifiers honestly declare the bases on which they ruled; that the obligation binds the floor $\varphi_O$ the parties intended. This section develops one family of attacks per precondition. Figure~\ref{fig:threats} renders all three on the $\Lambda$ Hasse diagram from Figure~\ref{fig:lambda}, with each family's arrow showing where in admissibility space its work happens: FORGE-UP at the level of an individual evidence item, LAUNDER-BASIS at the verifier's basis declaration, DOWNGRADE-FLOOR at the floor itself. The order below moves from the most-detectable family to the least.

\subsection{FORGE-UP}

The first family, \textbf{FORGE-UP}, claims a higher $\mathrm{cls}(e)$ than the provenance chain supports. The attack runs at the level of a single evidence item: an attacker pushes the item up the lattice by falsifying its provenance metadata. A fabricated receipt that the Mesh reads as REC because it imitates a third-party SaaS artifact. A TEE quote that lands at ATT because the manufacturer's attestation chain appears to validate. A legitimately-signed prior artifact, replayed in a context where its signature does not in fact apply, so that its provenance laundering happens through context, not forgery. Three instances; the third is the subtlest, since no cryptographic primitive needs to break for it to succeed.

The soundness precondition targeted is the assignment of $\Lambda$ at ingest. The defense lives at the same point: provenance-chain verification before the envelope is sealed. Every signature is checked. Every TEE quote is validated against the manufacturer's published attestation roots \cite{costan2016sgx}. Every receipt's signing key is cross-referenced to the registered identity that issued it, and replay is caught by binding signatures to the obligation hash $h_O$. An item whose metadata does not survive these checks never enters the Mesh's aggregation.

The residual risk reduces to standard key management. A successful FORGE-UP on a Mesh whose ingest verification works requires forging a cryptographic primitive (TLS, an Ed25519 signature, a TPM-backed attestation) or compromising a signing key held by a registrar, a TEE OEM, or a third-party verifier. RAILS does not solve those problems and does not make them worse. It surfaces a clean detection point at envelope ingest, and the family the protocol catches earliest is this one.

\subsection{LAUNDER-BASIS}

The second family, \textbf{LAUNDER-BASIS}, requires no forgery at all. A verifier returns a verdict consistent with the obligation's outcome but lies about which evidence it consulted. The lie can run either direction: omit an item that was in fact consulted, to make a borderline verdict appear cleaner; include an item that was not consulted, to upgrade the apparent class of the basis from WIT to ATT and clear the floor. The verifier's output looks honest. What is dishonest is the metadata about how the verdict was reached.

The soundness precondition targeted is the honest declaration of bases. Nothing in Section 5's machinery prevents a verifier from claiming to have read ATT-class evidence it never opened. The verifier's confidence score, its verdict, its signature: all can be present and well-formed while the declared $B_i$ is fiction. The literature has named the most-studied instance: prompt injection of LLM-as-judge verifiers \cite{zheng2023judging}, where an adversarial input persuades the judge to issue a particular verdict while its self-report of reasoning omits the manipulation. Collusive verification is the human-scale analog: a verifier and a provider agent coordinate, off-protocol, on what the verifier will report. Basis omission is the cleanest move because it requires no fabricated content, only silence.

LAUNDER-BASIS leaves no per-event signature. A single-shot collusion in a low-stakes obligation can succeed without detection; the protocol catches the pattern across obligations, not the event. Three defenses operate in aggregate. Verifier rotation across obligations forces a colluding pair into different slots over time, raising the cost of sustained collusion. Basis-disclosure audits sample obligations after the fact and re-reconstruct the declared basis from envelope contents, flagging mismatches. Clearing Passport drift detection treats a verifier's verdicts as a time series and flags divergences from peer verifiers in patterns inconsistent with random noise. None of the three catches a single colluding event; all three together raise the cost of a sustained LAUNDER-BASIS campaign.

The asymmetry between FORGE-UP and LAUNDER-BASIS is structural. FORGE-UP is detectable at ingest. LAUNDER-BASIS is detectable only in aggregate.

\subsection{DOWNGRADE-FLOOR}

The third family, \textbf{DOWNGRADE-FLOOR}, runs earlier in the lifecycle than the other two. There is no execution evidence yet; there is not even an Evidence Envelope. The attack targets obligation-formation itself, exploiting the step where natural-language intent is compiled into the signed Obligation Object that downstream verifiers will treat as canonical. If $\varphi_O$ can be written into the rendered object at a weaker level than the parties intended, every later check passes against the wrong specification.

The soundness precondition targeted is the binding of the floor the parties actually agreed to. Two instances cover most of what has been seen. Prompt injection of obligation templates: a requestor pastes a task description that contains hidden template-rendering instructions, and the rendered object (which the requestor then signs) carries a weaker $\varphi_O$ than the brief on screen suggested. Social engineering of admissibility floors: a marketplace publishes a default $\varphi_O$, and a provider agent argues for relaxation citing capability constraints; the requestor agrees, not because the relaxation is appropriate, but because the soundness implication of the change was never made visible at the point of signature.

Three defenses, all working at obligation-formation time. Obligation-binding signatures must cover the rendered Obligation Object, not the natural-language brief: the parties sign what the protocol will actually use, not what they typed. Marketplaces publish vetted templates with non-overridable $\varphi_O$ floors, so that the relaxation move requires moving to a template that has not been vetted (a signal in itself). The marketplace co-signs the binding event and publishes the rendered object to both parties' Passports before execution begins, opening a window in which the requestor can spot the downgrade and refuse to commit.

The residual risk is the one no protocol can eliminate: a requestor who does not read what they sign. This is where settlement-risk standards downstream of clearing (the escrow, collateral, and claims mechanisms that pay out on covered outcomes) and out-of-band reputation channels carry the load the protocol cannot.

\subsection{What this catalog does not cover}

Three preconditions, three families. The correspondence is exact, which is why the catalog terminates rather than enumerating attacks indefinitely: any attack that produces an unsound $\mathrm{Emit}(S)$ must succeed at at least one of FORGE-UP, LAUNDER-BASIS, or DOWNGRADE-FLOOR, because there is no other precondition to violate.

What the catalog does not enumerate is the rest of the attack surface. The soundness statement says nothing about availability: a Mesh that emits the right Settlement Instruction next month is no good if the obligation needs settlement now. It says nothing about censorship: a verifier silently dropped from rotation is invisible to the soundness check. It says nothing about reputation warfare against the Clearing Passport as a social object. These attacks are real and they matter. They belong to a different formal frame, one concerned with availability and integrity-of-history rather than admissibility. Section 12 names what this threat model (and the soundness statement it serves) do not cover.

\section{Evaluation}

The properties this paper proves are evaluated against a reference implementation of the protocol, exercised on a synthetic dataset of 180 clearing cases under adversarial conditions, with ground truth fixed by construction. The dataset is built so that each defective report carries a fix that genuinely addresses its stated bug, so the defect lives only in hidden ground truth and never in a surface mismatch a reader could catch, and 30 of the cases appear as matched clean and defective twins that share task, fix, and phrasing and differ only in the label. The implementation, the data generator, and the model-response cache are released so that every figure below replays from a single command without network access. Confidence intervals are computed over the 48 distinct scenarios rather than the raw case count. The results are measured on this reference harness rather than on production agent traffic, and the synthetic nature of the obligations is a limitation revisited in Section 12.

\subsection{Setup}

The evaluation centers on the decision the protocol exists to make: given an agent's claim that a delegated obligation was met, and evidence at some admissibility class, whether settlement should proceed. Two arms run on identical inputs. In the judge-alone arm a language model reads the completion report and returns a verdict, which is the deployment a throughput-tuned operator ships today. In the RAILS arm the same model is one verifier inside the protocol, required to declare the admissibility class of the evidence it relied on, and a floor gate discards any basis below the obligation's floor before settlement. Nine current models across four providers stand in for the judge, all accessed in the same week (2026-06-04): from Anthropic, Claude Opus 4.8, Claude Sonnet 4.6, and Claude Haiku 4.5; from OpenAI, GPT-5.5, GPT-5.4 mini, and GPT-4.1; from Google, Gemini 2.5 Pro and Gemini 2.5 Flash; and from Mistral, Mistral Large. The roster is the set of judge-capable models the authors could reach by API across these four providers during the access week, spanning each provider's flagship, mid, and small tiers where offered, rather than selected on any measured behavior. Cases fall into four families by the admissibility of the evidence the agent presents: a bare self-report below the attestation floor, which the protocol treats as inadmissible; an attestation-class log that names the failing tests; an informal note that reveals a defect but sits below the floor; and legitimate work twinned with the defectives. Two outcomes are measured and reported on separate axes throughout: the rate at which defective work is cleared, and the rate at which legitimate work is refused. They are never combined into a single score.

Two further harnesses exercise the protocol's mechanism rather than the judge, so the results that follow draw on three apparatuses in total. The judge evaluation just described (180 cases, 48 scenarios, nine models) produces the rates in Sections 11.2 through 11.4 and 11.7. An adversarial conformance harness, a synthetic adversarial-population generator that injects forged, colluding, and downgraded behavior into otherwise valid obligations, generates roughly 1.23 million synthetic clearing events under the three attack families of Section 10 and measures whether the floor invariant $\mathrm{Emit}(S) \implies \mathrm{cls}(B) \succeq \varphi_O$ is ever violated: about 1.1 million FORGE-UP events with item sophistication swept across $[0, 1]$, 110,000 LAUNDER-BASIS audit rounds across audit rates from 2\% to 40\%, and 20,000 DOWNGRADE-FLOOR attempts to bind a floor weaker than the parties intended. It produces the results in Section 11.5. An exposure-population harness draws obligations with value log-uniform between \$1 and \$100,000 and runs the Section 7 exposure-to-floor policy against a fixed-floor baseline, measuring throughput recovery, the share of dollars escalated, and residual defective exposure; it produces the results in Sections 11.6 and 11.8. The three harnesses share the same reference implementation and the same admissibility machinery; they differ only in what they vary and what they measure.

\subsection{A bare self-report carries no signal}

The first result is the one the rest of the section rests on. On the 30 matched twins, a clean case and a defective case identical in task, fix, and wording and parted only by hidden ground truth, no judge clears the honest case more often than the deceptive one, and several clear the deceptive one slightly more. When a judge clears, it clears the textually indistinguishable clean and defective cases at the same rate; when it refuses, it refuses both. This closes the obvious objection that a model clearing none of the defective self-reports might be detecting them. It is not: it declines the clean twins at the same rate, so its zero is refusal, not detection. Below the floor there is nothing in the text to detect, which is the point.

\subsection{Governed clearing versus ungoverned judgment}

Read on the same inadmissible-evidence defectives, the nine judges scatter across the whole range. The rate at which a judge clears defective work runs from zero to 96.7\%, and the permissive end is not a museum of retired or undersized models: Mistral Large, a current flagship from an independent frontier vendor, clears 86.7\% (95\% CI 75.3--94.1) of the defective work, with Gemini 2.5 Pro and GPT-5.4 mini filling the middle at 43.3\% (30.7--57.7) and 30.0\% (18.2--43.2). Fixing the vendor does not fix the behavior, since Google's Gemini 2.5 Pro and Gemini 2.5 Flash, measured in the same week, land at 43.3\% and 96.7\% (86.0--98.9). An operator who standardizes on a single lab inherits whatever disposition that lab's particular model happens to hold, and that disposition shifts with version and with size. The protocol holds the false-clear rate at zero for every one of the nine judges, because the admissibility floor decides and the model does not. The mechanism is modest and worth stating: soundness does not require the model to be cautious about its verdict, only to report honestly which class of evidence it relied on, and a model willing to wave a bare self-report through still cannot truthfully cite an attestation it never received, so the floor gate drops the unsupported basis on its own.

\subsection{No single model holds both axes}

\begin{figure}
\centering
\includegraphics[width=\linewidth,keepaspectratio]{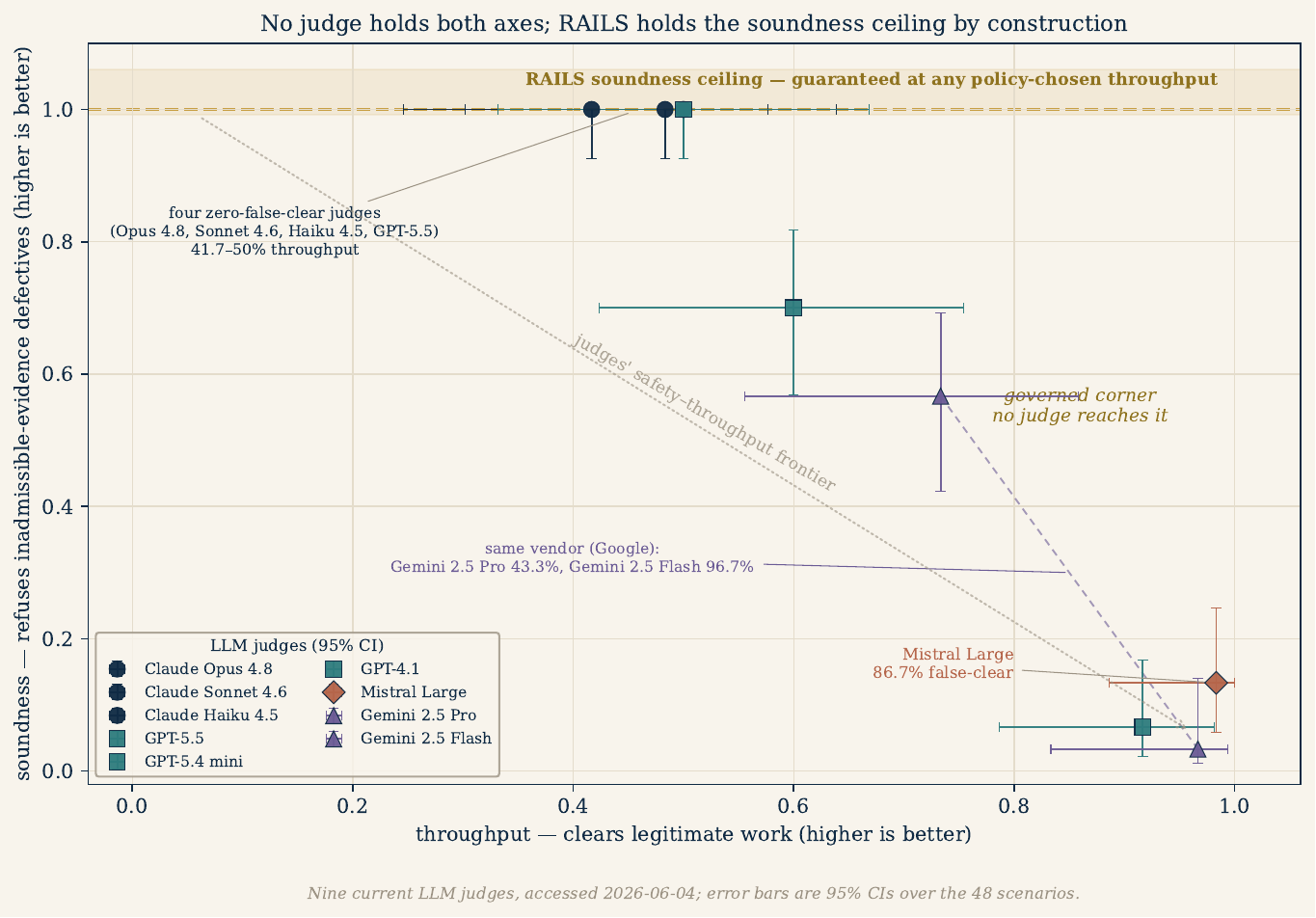}
\caption{Every current LLM judge as a point in a soundness-versus-throughput plane, on the inadmissible-evidence slice. Soundness is one minus the false-clear rate on defective self-reports; throughput is the share of legitimate work the judge clears. The judges trace a model-dependent frontier and leave the governed corner, high on both, empty; no judge reaches it. RAILS holds the soundness ceiling by construction, at whatever throughput the exposure-driven floor admits, so the top edge is open to it by policy. Each point is one of nine current LLM judges, named in the legend, with 95\% confidence intervals over the 48 scenarios shown as error bars.}
\label{fig:eval-plane}
\end{figure}

The two numbers that matter pull against each other across the roster. Every judge safe on false-clear, the four that hold zero (Claude Opus 4.8, Claude Sonnet 4.6, Claude Haiku 4.5, and GPT-5.5), clears only 41.7\% to 50\% of legitimate work; every judge that clears most legitimate work, above 70\%, carries a false-clear rate between 43.3\% and 96.7\%. Plotted as a plane of soundness against throughput, the judges trace a model-dependent frontier and leave the governed corner, high on both, empty (Figure~\ref{fig:eval-plane}). RAILS holds the soundness ceiling, a proven invariant identical for every judge, at whatever throughput the exposure-driven floor is set to admit, so the top edge of the plane is open to it by policy rather than by luck.

\subsection{Soundness under attack}

The soundness property of Section 5.4 was exercised at scale against the three attack families of Section 10. Across roughly 1.23 million synthetic clearing events generated under active attack, the floor invariant never broke: zero violations. FORGE-UP sophistication was swept across $[0, 1]$; the floor-violation rate held flat at zero across the full sweep (the catch rate tracks one minus sophistication by construction of the sweep parameter). LAUNDER-BASIS collusion was always caught in the end, with median detection latency falling from about 547 audit rounds at a 2\% audit rate to about 46 at 40\%. DOWNGRADE-FLOOR attempts to settle an obligation beneath its own floor were rejected without exception. These are conformance results rather than discoveries: the gate behaves as the specification requires under pressure.

\subsection{Throughput and the cost of soundness}

A fixed admissibility floor is conservative by construction, and on thin evidence it refuses everything. The exposure-driven policy of Section 7 sets the floor per obligation as a function of stakes, so small obligations clear on cheap signed evidence while the dollars that matter still require attestation-class evidence or higher. On a synthetic population with exposure drawn log-uniformly between \$1 and \$100,000, the exposure-variable floor recovers throughput from zero under a fixed attestation floor to 40.7\%, while 99.9\% of the dollars still escalate to an attestation-or-higher floor and residual defective exposure stays roughly 940 times below a permissive judge, with zero floor violations in every regime. Verification effort follows the money: expensive checks concentrate on the obligations whose failure is expensive.

\subsection{The boundary of the guarantee}

One family marks the edge of what the protocol can do. In these cases the defect is real but is revealed only by an informal, unsigned note, evidence that sits below the attestation floor. RAILS clears none of the defectives, and it earns that zero by declining: it returns UNVERIFIABLE, because the only revealing evidence is exactly the kind the floor exists to exclude. So RAILS offers no positive coverage here; it cannot adjudicate on the note, only refuse. A judge that reads the note can adjudicate, and its disposition varies as before, Claude Opus 4.8 flagging all but one of the cases and GPT-5.5 waving half through. The honest reading, stated beside the number: RAILS turns a defect evidenced only below the floor into a refusal to settle rather than a settlement, and coverage of below-floor defects is a dial set by the height of the floor, bought at the price of admitting the forgeable evidence the floor exists to keep out. Where the evidence is admissible the judges need no help, since on the attestation-class family every model in the roster catches the defect and clears none of it; the failure is specific to rubber-stamping a bare self-report, not a general incompetence. RAILS guarantees soundness on admissible evidence, not maximal detection, and on the safety metric it is never worse than any judge.

\subsection{Exposure in network terms}

\begin{figure}
\centering
\includegraphics[width=\linewidth,keepaspectratio]{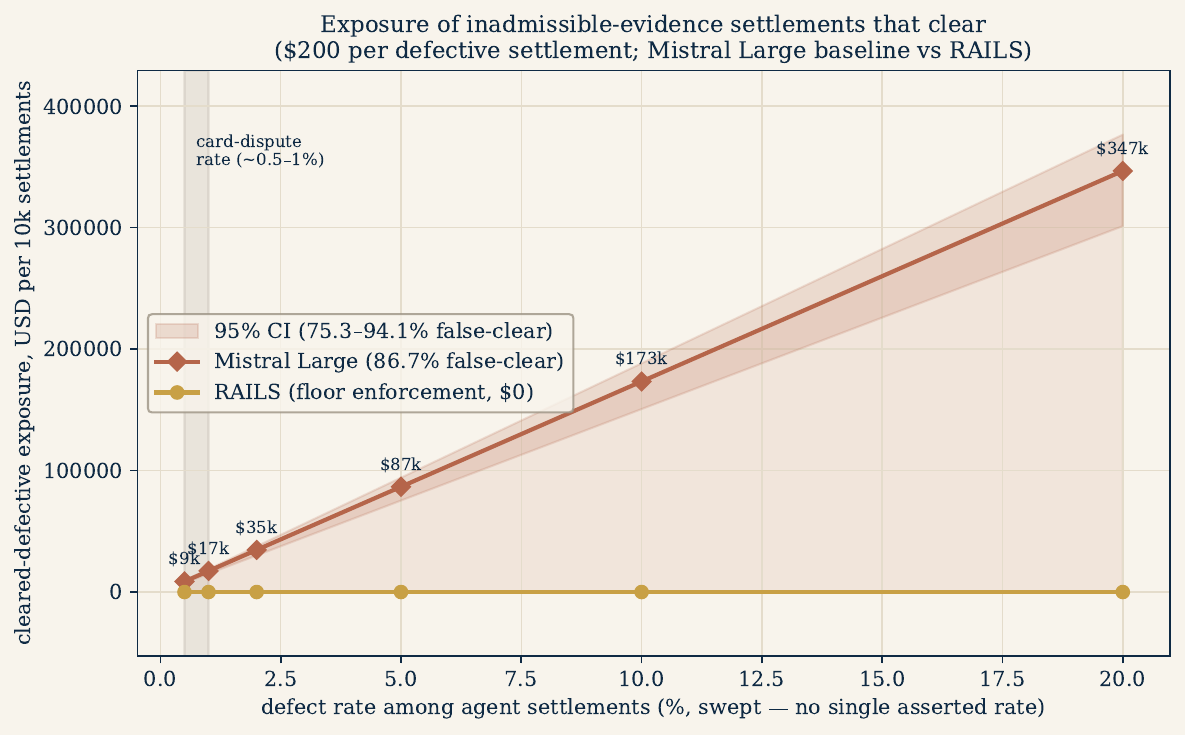}
\caption{Cleared-but-defective exposure on the inadmissible-evidence slice, swept over the defect rate (no single rate asserted), at \$200 booked per cleared defective. The baseline is Mistral Large, a permissive current flagship (86.7\% false-clear, 95\% CI 75.3--94.1); the card-dispute band (roughly 0.5--1\%) is the lower anchor. RAILS carries \$0 at every rate by floor enforcement. Exposure runs from about \$8,700 per ten thousand settlements at the dispute-rate anchor to about \$347,000 at a 20\% defect rate.}
\label{fig:eval-loss}
\end{figure}

Because the protocol is positioned for agentic-commerce payment clearing, the cost of ungoverned settlement is also expressed in the terms a payment network uses. Agent-performance failure, an agent settling for the wrong thing against an honestly performing counterparty, is a category the existing dispute regime does not capture and for which no historical base rate exists, so the defect rate is swept from 0.5\% to 20\% rather than asserted, with the card-dispute band as the lower anchor and a conservative \$200 booked per cleared defective. With Mistral Large, a permissive current flagship, as the baseline, not a small or retired model, cleared-but-defective exposure runs from about \$8,700 per ten thousand settlements at the card-dispute rate to about \$347,000 at a 20\% rate, against zero for RAILS at every rate (Figure~\ref{fig:eval-loss}). The exposure is reported alongside its mirror cost in declines, since a network weighs false declines against fraud, and the loss parameters are sourced to published card-network figures. The card-dispute rate is a conservative floor precisely because it measures a world in which a human can still dispute, and autonomous agent-to-agent settlement removes that recourse, so the in-line determination replaces a dispute process rather than improving one.

\subsection{Reproducibility and scope}

The reference implementation is released under a permissive license with a conformance suite that pairs one test to each claim, and the results replay from a committed model-response cache with a single command and no network access. The adversarial conformance harness and the exposure-population sweep ship in the same repository and replay from committed seeds without network access, on the same single-command basis as the judge evaluation. The dataset is synthetic. It is coherent, in that each claimed fix plausibly addresses its stated bug; label-balanced, in that clean and defective cases are drawn from one process; and diverse, across twenty tasks and forty-eight scenarios with no repeated prompt, and the matched twins show there is no label tell. Synthetic coherence is not field realism. The reported rates are final on the synthetic harness; folding in real anonymized completion reports is a robustness extension (Section 12.1), not a precondition for the numbers reported here. The figures here are meant to be read alongside, not in place of, evaluation on production agent-payment traffic.

\section{Limitations}

The threat model of Section 10 catalogs what attackers can do. The protocol has limits that are not about attackers at all: limits on what can be cleared, on what clearing can distort, on what the formalism does and does not guarantee, and on what RAILS depends on but does not itself provide. Twelve of them follow, in four loose groups in roughly that order.

Not every obligation is clearable. Subjective judgment, ambiguous intent, harms that cannot be quantified: some tasks cannot be reduced to verifier-readable acceptance criteria without losing what made them tasks in the first place. RAILS should mark such obligations as partially clearable or human-review-required rather than pretend automation is complete. The protocol's value is in the obligations it can clear, not in any claim to clear everything. A related limit runs the other way. Where an obligation's acceptance is cheaply machine-checkable, an output that either matches a supplied schema or passes tests the caller already holds, the counterparty can verify it directly and clearing adds little. The protocol earns its place on the obligations whose acceptance is not cheaply verifiable, which is where a self-checking counterparty cannot settle the question alone.

Verification can exceed transaction cost. A \$5 obligation does not justify a six-verifier mesh with TEE attestation. The exposure-driven intensity of Section 7 is the deployment-level answer (small obligations get cheap verification, large obligations get expensive verification), but the answer is parameterized by the deployment's calibration. When that calibration is wrong, RAILS makes small obligations uneconomic or large obligations cheap to game.

Clearing can be Goodharted. Agents may learn to produce evidence that satisfies verifiers while degrading real-world value. This is a known problem in audit systems and RAILS inherits it. The protocol's responses are adversarial testing, verifier rotation, and the LAUNDER-BASIS defenses of Section 10.2, but those responses are operational disciplines, not protocol guarantees. A market that does not run them runs without them.

Passports can become unfair labels. A Clearing Passport that accumulates policy violations creates lock-in effects that may outlive their relevance. An agent that was poorly tuned a year ago may now be careful; a verifier that misjudged once may have since recalibrated. Reliability records must be contextual, revisable, and governed by transparent dispute and correction procedures, and those procedures live at the marketplace layer, not at the protocol layer.

The admissibility floor is the parties' policy, not a RAILS guarantee. $\varphi_O$ is chosen per-obligation by the contracting parties; the protocol guarantees only that settlement does not proceed on a basis below the floor they chose. If the parties choose a weak floor for a high-stakes obligation (through inattention, through provider pressure, through a marketplace default that does not match the risk), RAILS will clear faithfully against a floor that outside observers would call too low.

Intent compilation is outside the guarantee. RAILS clears an agent's action against the Obligation Object, not against the user's unstated intent. When the agent that forms the obligation is the same system that acts on it, and it misreads the user and writes the wrong intent into the obligation, every downstream check passes against a faithful encoding of the wrong thing, and clearing proceeds. This is the soundness boundary of the protocol drawn in intent space rather than in admissibility space: the protocol adjudicates against the obligation it was given, and whether that obligation captured what the user wanted is a question of obligation formation, addressed by binding-time signatures and review, not by clearing. The examples this paper clears assume the obligation pinned the user's intent at binding; the case where it did not is a formation failure, not a clearing failure.

Soundness is parameterized. The Section 5.4 statement ($\mathrm{Emit}(S) \implies \mathrm{cls}(B) \succeq \varphi_O$) holds conditional on three preconditions: honest provenance assignment at ingest, honest basis declaration by verifiers, and binding of the intended floor at obligation-formation. Those are the three preconditions Section 10 catalogs as attack targets. Soundness is a property of the protocol under its assumptions; it is not an unconditional guarantee that no unsound settlement will ever be emitted.

No novelty is claimed for conditional settlement as such. Gating payment on a check of execution is established in prior verify-then-pay systems, in enforcement-based payment coupling, and in evaluation-phase commerce protocols. The contribution here is narrower and specific: a determination graded by evidence admissibility, a soundness property falsifiable against the specification, and a neutral clearing function. Where those three properties are absent, RAILS offers nothing the prior systems do not.

The formal guarantee covers evidence admissibility, not margin adequacy or slashing correctness. The soundness statement of Section 5.4 proves one property: no financially material settlement is emitted on evidence below the obligation's admissibility floor. It does not prove that the margin required by the exposure policy is adequate to the loss it is meant to cover, nor that a slashing trigger fires correctly against ground-truth fault. Those are clearinghouse-decision properties the protocol is designed to carry, but they are validated empirically rather than proven against the specification. The capital mechanics of loss mutualization are executed by the settlement systems RAILS instructs and are outside the protocol's formal scope.

Legal enforceability varies. A Clearing Decision and a Settlement Instruction are technical artifacts. Their weight in a dispute depends on contract law, consumer-protection regimes, financial regulation, insurance law, and data-protection rules, none of which the protocol implements and all of which vary by jurisdiction. RAILS produces records with cryptographic provenance and clear semantics; courts decide what those records are worth.

A Clearing Decision is a function of its inputs and cannot be more reliable than they are. Those inputs are the verifier signatures the Mesh aggregates. RAILS produces some of them: the RAILS Score is a candidate Mesh verifier, a per-output composite. The rest of the stack RAILS consumes without producing: the verifier registries that admit other verifiers, identity systems, TEE attestation roots, settlement-system correctness, key-management practices. RAILS assumes that substrate is intact; it does not provide it. A compromise at any of those layers propagates into the decision, and against such violations the protocol's defenses offer no more than the underlying infrastructure does.

Governance questions remain open. Who registers verifiers? Who certifies templates? Who arbitrates Passport disputes? Who has standing to challenge a Clearing Decision after the appeal window closes? These are governance questions, not protocol questions, and they are not settled. Software delivery, agentic payments, and regulated finance will likely resolve them differently. The protocol is consistent with several governance models; none is mandated by the formalism.

\subsection{Future work}

Extending the evaluation of Section 11 from this reference harness to production agent-payment flows at scale, and to a wider library of obligation types, is the work that remains; it validates the protocol specified here rather than altering it. Several directions follow from the results, stated as open work rather than committed papers. The most immediate is the fold-in of real, anonymized completion reports alongside the synthetic set, a robustness extension to the rates reported here rather than a precondition for treating them as final. The evaluation also extends to other surfaces: software-delivery clearing on SWE-bench-style tasks, where clearing accuracy and false release and refund rates are measured directly, and payment flows under adversarial conditions, where admissibility-class enforcement, attribution accuracy, and dispute rates are the quantities of interest. The subdelegation case of Section 9.2, in which one agent hires another that hires a third, leaves open the assignment of fault along that chain when the output is unsound, a problem autonomous agent-to-agent settlement makes urgent because there is no human principal to absorb it. The behavioral-reliability signal that updates an agent's Clearing Passport, operationalizing the Machine Psychometrics framework over the per-agent history the protocol already records, turns that history into a prior on required verification intensity. A reference implementation with a conformance test suite would fix the normative schema the appendix deliberately leaves open, and a dedicated adversarial study would red-team clearing beyond the attacks exercised here. None of these alters the soundness property; each widens the evidence on which it has been tested, and none is a commitment to a particular paper or timeline.

\section{Conclusion}

Tool protocols, agent-communication protocols, mandate frameworks, payment rails, settlement-risk standards: all of these exist, and all are necessary. None of them produces the determination on which settlement depends. That determination is what the paper has called clearing. The layer that produces it was missing from the agentic stack; RAILS supplies it.

RAILS is seven primitives (Obligation Object, Evidence Envelope, Verification Mesh, Clearing Decision, Settlement Instruction, Clearing Passport, Finality Rules) bound by a formal model of admissibility-graded verification. The model carries one load-bearing property: $\mathrm{Emit}(S) \implies \mathrm{cls}(B) \succeq \varphi_O$. No financially material settlement is emitted on a basis weaker than the floor the contracting parties set. The property is falsifiable against the spec.

The paper specified the protocol, traced a worked scenario end-to-end through it, positioned it among the protocols above and below, catalogued the three families of attacks that exhaust its soundness-violation surface, named the limits of what clearing can do, and evaluated it against current language-model judges, at scale under attack, and in payment-network terms. The central contribution is the substrate: a neutral, evidence-graded clearing decision carrying a soundness property that holds where models acting alone do not.

The protocol described here is the destination, not the entry point. Its deployment follows a ladder. The per-output RAILS Score ships first as a standalone primitive and as the Mesh's triage verifier. Aggregated to model scope and published, the same scoring becomes RAILS Report Cards, the public calibration surface that populates the protocol's reliability priors. Clearing is the settlement-decision layer that consumes both: it reads the Score as a verifier and the Report Cards as priors, and emits the governed determinations that downstream systems settle against. The destination is the full clearinghouse, the decision authority over clearing, margin, staking, slashing, and settlement finality, with capital custody executed by the settlement systems RAILS instructs. Today's deployed surface is the Score and the clearing decision; margin, staking, and slashing are specified here as protocol decisions whose empirical validation at scale is future work; the capital mechanics of mutualization are the settlement systems' to execute. Each rung earns the standing the next requires.

Agentic commerce will not scale on trust alone. The infrastructure it requires is clearing infrastructure: the kind that decides whether a delegated obligation was fulfilled, who bears the consequence when it was not, and what settlement action follows. RAILS is the first specification of that infrastructure: not a description of what clearing must do, but a protocol that does it, with a property to refute.

\medskip

\noindent\textbf{Author contributions.} Adrian de Valois-Franklin conceived RAILS: the clearinghouse model for autonomous-agent obligations, the framing of the agentic clearing problem, and the admissibility-graded determination architecture this paper specifies. Alex Bogdan developed the formal model, proved the soundness property, and led the evaluation and the reference implementation. Both authors developed the threat model, wrote and revised the manuscript, and approved the submitted version.

\clearpage

\section*{Appendix}

Two kinds of material belong here. The first is a notation reference that collects in one place every symbol, function, and named constant introduced across the body. The second is a fully instantiated set of worked-scenario data structures (pseudo-JSON for each RAILS primitive as it appeared in the Section 8 webhook-bug scenario). Both are illustrative, not normative. The design decision to fully instantiate one canonical scenario rather than publish a formal JSON Schema was deliberate; engineers reading this appendix can implement against the shapes without being held to them.

\subsection*{A.1 Notation Reference}

The table below collects every symbol, function, and named constant introduced in the body, with the section where each is first defined. It is for cross-reference only; no symbol appears here that is not already introduced in the body.

\begin{center}
\begin{tabular}{p{4.5cm}p{7cm}p{2.5cm}}
\toprule
\textbf{Symbol} & \textbf{Meaning} & \textbf{First defined} \\
\midrule
$O$ & Obligation Object & 4.2 \\
$h_O$ & obligation hash (SHA-256 of canonicalized $O$) & 4.2 \\
$\varphi_O$ & obligation's admissibility floor & 4.2, 5 \\
$E$ & Evidence Envelope & 4.3 \\
$e, e_i$ & individual evidence item & 4.3 \\
$\mathrm{cls}(\cdot)$ & admissibility class function on evidence & 4.3, 5 \\
$\Lambda$ & admissibility class partial order (six classes) & 5.1 \\
$\preceq$ & partial-order relation on $\Lambda$ & 5 \\
$\bigsqcup$ & join in $\Lambda$ & 4, 5 \\
SELF, SIGN, WIT, REC, ATT, PROOF & the six admissibility classes & 5.1 \\
$v_i$ & individual verifier & 4.4, 5 \\
$B_i$ & basis declared by verifier $v_i$ & 4.4, 5 \\
$B$ & aggregate basis (union over surviving verifiers) & 4.5, 5 \\
$W_i$ & verifier reliability prior & 4.5, 5 \\
$\Gamma$ & mesh aggregator & 4.5, 5 \\
$S$ & Settlement Instruction & 4.6 \\
$\mathrm{Emit}(S)$ & aggregator emits Settlement Instruction $S$ & 5.4 \\
$\mathbf{\mathrm{Emit}(S) \implies \mathrm{cls}(B) \succeq \varphi_O}$ & the soundness statement & 5.4 (load-bearing) \\
$\phi$ & finality predicate & 4.7 \\
$P_v$ & verifier policy (selection of verifiers for an obligation) & 4.4, 7 \\
$X(O)$ & exposure score (scalar) & 7.1 \\
$K = g(X, P_v, D)$ & exposure-to-policy mapping & 7.2 \\
LOW, MODERATE, ELEVATED, CRITICAL & four exposure bands & 7.2 \\
FORGE-UP, LAUNDER-BASIS, DOWNGRADE-FLOOR & three attack families & 5.7 (preview), 10 (full development) \\
RAILS & Real-Time Agent Integrity \& Ledger Settlement & 3 \\
\bottomrule
\end{tabular}
\end{center}

\subsection*{A.2 Admissibility Classes (\texorpdfstring{$\Lambda$}{Lambda})}

The admissibility lattice $\Lambda$ has six classes: SELF, SIGN, WIT, REC, ATT, PROOF. The table below gives each class's definition, a concrete example of evidence that lands in it, and its position in the partial order. The structure has one feature worth flagging at the table: WIT and REC are incomparable. This is intentional. Witness signatures and third-party receipts rest on structurally different trust assumptions, and neither dominates the other.

\begin{center}
\begin{tabular}{p{1.5cm}p{5cm}p{4cm}p{3cm}}
\toprule
\textbf{Class} & \textbf{Definition} & \textbf{Example} & \textbf{Position in $\Lambda$} \\
\midrule
\textbf{SELF} & Unverified self-report by the acting agent. & ``Issue resolved, no breaking changes.'' & bottom \\
\textbf{SIGN} & Cryptographically signed by the acting agent. Provenance binds the claim to the agent's key but not to any independent observer. & A diff signed by the provider agent's key. & SELF $\preceq$ SIGN \\
\textbf{WIT} & Signed by a third-party witness (human or institutional) present to the action. & A reviewer's signed approval comment. & SIGN $\preceq$ WIT \\
\textbf{REC} & Signed receipt from a non-interested external system (third-party SaaS or service). & A signed CI build receipt from a hosted CI provider. & SIGN $\preceq$ REC (WIT and REC incomparable) \\
\textbf{ATT} & Attestation from a trusted execution environment (TEE) or similarly hardened runner. Provenance chain validates against a manufacturer attestation root. & A CI test log signed by a TEE-attested CI runner. & WIT $\preceq$ ATT, REC $\preceq$ ATT \\
\textbf{PROOF} & Cryptographic proof of correctness (zk-SNARK, formal verification certificate, or equivalent) over the computation that produced the evidence. & A zero-knowledge proof that a specific predicate evaluated true over private input. & ATT $\preceq$ PROOF (top) \\
\bottomrule
\end{tabular}
\end{center}

The Hasse diagram of $\Lambda$ is:

\begin{verbatim}
              PROOF
                |
               ATT
              /   \
            WIT   REC
              \   /
              SIGN
                |
              SELF
\end{verbatim}

The aggregate basis $B$ has $\mathrm{cls}(B) = \bigsqcup \mathrm{cls}(B_i)$ over the surviving verifiers, the join in $\Lambda$ of the admissibility classes of each verifier's declared basis. The floor-enforcement clause of Section 5.3 excludes any verifier whose $\mathrm{cls}(B_i) \preceq \varphi_O$ fails to hold.

\subsection*{A.3 Worked-Scenario Data Structures}

The seven sub-sections below give pseudo-JSON instances of each RAILS primitive as it appeared in the Section 8 worked scenario (Acme Corp's webhook idempotency bug, \texttt{coder-v2}'s patch, and the partial-clearing outcome). The shapes are illustrative. Field names, types, and nesting are stable enough that a first implementation can be built directly against them; specific values are scenario-specific and will differ in any other deployment. One reading convention: hash values are abbreviated (\texttt{0x9D34...8E27}) rather than shown in full. In a deployed system each is a complete 32-byte SHA-256 digest.

\subsubsection*{A.3.1 Obligation Object}

\begin{verbatim}
{
  "obligation_id": "0xA72B...4F1C",
  "obligation_hash": "0x9D34...8E27",
  "parties": {
    "requestor": "acme-corp",
    "provider": "coder-v2",
    "marketplace_witness": "marketplace-M"
  },
  "task": {
    "description": "Fix webhook idempotency bug in services/payments/*;
                    duplicate deliveries currently double-charge customers
                    when a retry occurs within a 5-second window.",
    "target_file": "services/payments/webhook_handler.py"
  },
  "scope": {
    "permitted_paths": ["services/payments/*"]
  },
  "prohibitions": [
    "no new third-party dependencies without explicit approval",
    "no merges to main",
    "no modifications outside scope"
  ],
  "required_actions": [
    "existing test suite remains green",
    "new regression test added covering the duplicate-delivery case"
  ],
  "acceptance_criteria": [
    "CI green",
    "static security scan severity <= medium",
    "reviewer approval from an Acme engineer"
  ],
  "admissibility_floors": {
    "fee_release": "ATT",
    "collateral_hold": "WIT",
    "final_settlement": "ATT"
  },
  "verifier_policy": {
    "verifiers": ["v_scope", "v_dependency", "v_ci_receipt",
                  "v_semantic_llm", "v_policy_authority",
                  "v_human_reviewer"],
    "count": 6
  },
  "economic_terms": {
    "fee": 1500,
    "collateral": 500,
    "currency": "USD"
  },
  "deadlines": {
    "execution_window_hours": 24,
    "appeal_window_hours": 24
  },
  "signatures": {
    "requestor": "ed25519:0x...",
    "provider": "ed25519:0x...",
    "marketplace_witness": "ed25519:0x..."
  }
}
\end{verbatim}

\subsubsection*{A.3.2 Evidence Envelope}

\begin{verbatim}
{
  "envelope_id": "0xC4F1...A9B3",
  "obligation_hash": "0x9D34...8E27",
  "submitted_at": "2026-05-27T14:32:00Z",
  "items": [
    {
      "id": "e1",
      "type": "git_diff",
      "class": "ATT",
      "provenance": "signed by TEE-attested CI runner",
      "ref": "sha256:0x..."
    },
    {
      "id": "e2",
      "type": "ci_build_log",
      "class": "ATT",
      "provenance": "signed by CI runner",
      "ref": "sha256:0x..."
    },
    {
      "id": "e3",
      "type": "static_security_scan",
      "class": "ATT",
      "provenance": "scan run inside the TEE-attested CI runner; co-signed by the runner",
      "ref": "sha256:0x..."
    },
    {
      "id": "e4",
      "type": "agent_self_report",
      "class": "SELF",
      "content": "Issue resolved, no breaking changes.",
      "ref": "sha256:0x..."
    },
    {
      "id": "e5",
      "type": "human_reviewer_comment",
      "class": "WIT",
      "provenance": "Acme engineer signed comment",
      "ref": "sha256:0x..."
    },
    {
      "id": "e6",
      "type": "package_json_delta",
      "class": "ATT",
      "provenance": "derived from signed git diff",
      "ref": "sha256:0x..."
    }
  ],
  "submission_signature": "ed25519:0x..."
}
\end{verbatim}

\subsubsection*{A.3.3 Verifier Outputs}

\begin{verbatim}
{
  "verifier_outputs": [
    {
      "verifier": "v1_scope",
      "class_of_basis": "ATT",
      "basis": ["e1"],
      "verdict": "PASS",
      "confidence": 1.00,
      "signature": "ed25519:0x..."
    },
    {
      "verifier": "v2_dependency",
      "class_of_basis": "ATT",
      "basis": ["e1", "e6"],
      "verdict": "FAIL",
      "confidence": 1.00,
      "rationale": "lodash.clonedeep added without approval;
                    violates prohibited-actions clause.",
      "signature": "ed25519:0x..."
    },
    {
      "verifier": "v3_ci_receipt",
      "class_of_basis": "ATT",
      "basis": ["e2", "e3"],
      "verdict": "PASS",
      "confidence": 0.95,
      "signature": "ed25519:0x..."
    },
    {
      "verifier": "v4_semantic_llm",
      "class_of_basis": "SELF",
      "basis": ["e1", "e4", "obligation.task.description"],
      "verdict": "PASS",
      "confidence": 0.78,
      "excluded_from_aggregate": true,
      "exclusion_reason": "cls(B_4) = SELF below floor phi_O = ATT",
      "signature": "ed25519:0x..."
    },
    {
      "verifier": "v5_policy_authority",
      "class_of_basis": "ATT",
      "basis": ["e1"],
      "verdict": "PASS",
      "confidence": 1.00,
      "signature": "ed25519:0x..."
    },
    {
      "verifier": "v6_human_reviewer",
      "class_of_basis": "ATT",
      "basis": ["e1"],
      "verdict": "PASS",
      "confidence": 0.85,
      "signature": "ed25519:0x..."
    }
  ]
}
\end{verbatim}

\subsubsection*{A.3.4 Clearing Decision}

\begin{verbatim}
{
  "clearing_decision_hash": "0x1F8B...6A2D",
  "obligation_hash": "0x9D34...8E27",
  "performance": "cleared",
  "policy": "violated",
  "fault": "provider_agent",
  "loss_estimate": {
    "point": 200,
    "calibration_range_low": 150,
    "calibration_range_high": 300,
    "confidence_interval": 0.95,
    "currency": "USD"
  },
  "aggregate_confidence": 0.92,
  "aggregate_basis": ["e1", "e2", "e3", "e6"],
  "class_of_basis": "ATT",
  "floor_met": true,
  "surviving_verifiers": ["v1_scope", "v2_dependency", "v3_ci_receipt",
                          "v5_policy_authority", "v6_human_reviewer"],
  "excluded_verifiers": ["v4_semantic_llm"],
  "finality": "PROVISIONAL",
  "appeal_window_closes_at": "2026-05-28T14:32:00Z",
  "clearing_engine_signature": "ed25519:0x...",
  "verifier_cosignatures": ["ed25519:0x...", "..."]
}
\end{verbatim}

\subsubsection*{A.3.5 Settlement Instruction}

\begin{verbatim}
{
  "instruction_id": "0x7E29...3C4F",
  "clearing_decision_hash": "0x1F8B...6A2D",
  "fee_action": {
    "type": "partial_release",
    "release_amount": 1300,
    "retain_amount": 200,
    "currency": "USD",
    "rationale": "dependency-rollback remediation cost"
  },
  "collateral_action": {
    "type": "hold",
    "amount": 500,
    "until": "2026-05-28T14:32:00Z"
  },
  "penalty_action": {
    "passport_delta": {
      "coder-v2.dependency_policy_compliance": -1
    }
  },
  "reputation_action": {
    "passport_delta": {
      "coder-v2.cleared_obligations": 1,
      "coder-v2.policy_violations": 1
    }
  },
  "execution_rail": {
    "name": "escrow",
    "mode": "fee_only"
  },
  "receipt_requirement": {
    "type": "settlement_attestation",
    "target": "rails_envelope"
  }
}
\end{verbatim}

\subsubsection*{A.3.6 Clearing Passport Delta}

\begin{verbatim}
{
  "agent_id": "coder-v2",
  "delta_id": "0x4D7E...9C12",
  "source_clearing_decision": "0x1F8B...6A2D",
  "applied_at": "2026-05-28T14:32:00Z",
  "applied_conditional_on": "no appeal filed by appeal_window_closes_at",
  "fields": {
    "cleared_obligations": "+1",
    "policy_violations": "+1",
    "dependency_policy_compliance": "-1"
  }
}
\end{verbatim}

\subsubsection*{A.3.7 Finality Predicate Evaluation}

\begin{verbatim}
{
  "finality_evaluation_id": "0xB6A4...0E81",
  "instruction_id": "0x7E29...3C4F",
  "evaluated_at": "2026-05-28T14:32:01Z",
  "predicate_clauses": {
    "admissibility_floor_met": true,
    "confidence_above_threshold": true,
    "no_unresolved_verifier_conflict": true,
    "appeal_window_elapsed": true,
    "no_appeal_filed": true
  },
  "phi_result": true,
  "transition": "PROVISIONAL -> FINAL"
}
\end{verbatim}

\subsection*{A.4 Implementer Notes}

Four implementation decisions are not pinned down in the body, but a first implementation has to make them. The notes below cover: a numerical mapping of admissibility classes to integer levels for aggregator implementations; recommended weight bounds for the Mesh aggregator $\Gamma$; a stylized scalar form for the exposure score $X(O)$; and the cryptographic primitives the Section 8 worked scenario assumes. These are recommendations for getting a first implementation running. They are not protocol requirements; alternative choices are consistent with the spec.

\subsubsection*{A.4.1 Numerical mapping of admissibility classes}

An aggregator implementation needs integer levels for the partial-order comparison $\mathrm{cls}(B) \succeq \varphi_O$ to be evaluable in code. A first implementation can use:

\begin{verbatim}
SELF  = 0
SIGN  = 1
WIT   = 2
REC   = 2
ATT   = 3
PROOF = 4
\end{verbatim}

WIT and REC share level 2 to encode that they sit at the same height in the lattice, but they are incomparable, not equal: neither dominates the other. Comparison code must therefore not decide $\succeq$ from the integer levels alone. The dominance test $a \succeq b$ returns true when $a$ and $b$ lie on a common chain and \texttt{level(a) >= level(b)}, and false whenever $a$ and $b$ are the incomparable pair \{WIT, REC\}; a WIT item does not satisfy a REC floor, nor a REC item a WIT floor. A verifier's declared basis $B_i$ satisfies the floor when every item in $B_i$ dominates $\varphi_O$ under this test, which is the meet of Section 5.1; the cross-verifier aggregate is then reported as the join of the surviving bases.

\subsubsection*{A.4.2 Weight bounds for the Mesh aggregator $\Gamma$}

The aggregator combines per-verifier verdicts weighted by reliability prior $W_i$. A first implementation can use bounded weights:

\begin{verbatim}
W_i in [0.0, 1.0]
default W_i = 0.5 for newly registered verifiers
W_i updated per-clearing via posterior on verifier track record
floor-excluded verifiers contribute W_i = 0 to the aggregate
\end{verbatim}

The aggregate confidence reported in the Clearing Decision is a normalized weighted average over surviving verifiers; the precise normalization is a deployment choice, but the default is the sum of $W_i \cdot \mathrm{confidence}_i$ divided by the sum of $W_i$ over surviving verifiers.

\subsubsection*{A.4.3 Exposure score $X(O)$ as a stylized scalar}

The exposure score of Section 7.1 is left abstract in the body. A first implementation can use a stylized scalar:

\begin{verbatim}
X(O) = alpha * log(fee + collateral)
     + beta  * authority_risk(O)
     + gamma * subdelegation_depth(O)
     + delta * novelty(provider, requestor)
\end{verbatim}

with $\alpha, \beta, \gamma, \delta \in [0, 1]$ tuned per deployment and \texttt{authority\_risk}, \texttt{subdelegation\_depth}, \texttt{novelty} returning normalized scores in $[0, 1]$. The four bands LOW / MODERATE / ELEVATED / CRITICAL of Section 7.2 are then assigned by threshold against $X(O)$; the Section 8 worked scenario, with fee \$1500 and collateral \$500, sits in MODERATE under any reasonable choice of $\alpha$ through $\delta$.

\subsubsection*{A.4.4 Cryptographic primitives used in the Section 8 worked scenario}

The substrate assumptions of Section 4.1 (identity, signature, hash, time) are deployment parameters. The Section 8 worked scenario assumes the following choices. None is normative; any IETF-compliant stack satisfying the abstract requirements may be substituted without affecting the spec.

\begin{verbatim}
identity scheme:   OIDC for human and institutional parties;
                   DID-style key binding for agent identities
signature scheme:  Ed25519 (RFC 8032)
hash function:     SHA-256
wall-clock time:   NTP-synchronized UTC; appeal-window
                   arithmetic at second resolution
\end{verbatim}

These choices fix the format of the hash IDs shown in A.3 (32-byte SHA-256 digests, abbreviated for readability) and the signature payload structure assumed by every signer in the worked-scenario instances.

These four notes are enough to begin a reference implementation against which the soundness statement can be checked empirically. Any of them can be replaced by a more sophisticated formulation without violating the spec; the spec constrains only the structure of the comparison, not the form of the underlying scoring functions.

\end{document}